\pdfoutput=1

\documentclass[11pt]{article}

\usepackage[final]{acl}

\usepackage{times}
\usepackage{comment}
\usepackage{latexsym}

\usepackage[T1]{fontenc}

\usepackage[utf8]{inputenc}

\usepackage{microtype}

\usepackage{inconsolata}

\PassOptionsToPackage{hyphens}{url}\usepackage{hyperref}
\usepackage{ulem}
\usepackage{booktabs}
\usepackage{enumitem}
\usepackage{newfloat}
\usepackage{mathtools}
\usepackage{listings}
\usepackage{multirow}
\usepackage{float}
\usepackage{caption}
\usepackage{subfigure}
\usepackage{graphicx}
\usepackage{stfloats}
\usepackage{xcolor}
\usepackage{fontawesome}
\usepackage{amsmath}

\usepackage{amsfonts}
\usepackage{amssymb}
\usepackage{algorithm}
\usepackage{breqn}

\usepackage{amsmath}

\usepackage{algpseudocode}

\title{Reinforcement Tuning for Detecting Stances and Debunking Rumors Jointly with Large Language Models}

\author{Ruichao Yang$^{1}$, Wei Gao$^{2}$, Jing Ma$^{1}$\thanks{\; Jing Ma is the corresponding author.}, Hongzhan Lin$^{1}$, Bo Wang$^{3}$\\
$^{1}$Hong Kong Baptist University, Hong Kong SAR, China\\
$^{2}$Singapore Management University, Singapore\\
$^{3}$Jilin University, Changchun, Jilin, China\\
\texttt{\{csrcyang,majing,cshzlin\}@comp.hkbu.edu.hk}, \\\texttt{weigao@smu.edu.sg}, \texttt{wangbo21@mails.jlu.edu.cn}
}

\begin{document}
\maketitle
\begin{abstract}
Learning multi-task models for jointly detecting stance and verifying rumors poses challenges due to the need for training data of stance at post level and rumor veracity at claim level, which are difficult to obtain. To address this issue, we leverage large language models (LLMs) as the foundation annotators for the joint stance detection (SD) and rumor verification (RV) tasks, dubbed as JSDRV. We introduce a novel reinforcement tuning framework to enhance the joint predictive capabilities of LLM-based SD and RV components. Specifically, we devise a policy for selecting LLM-annotated data at the two levels, employing a hybrid reward mechanism to choose high-quality labels for effective LLM fine-tuning on both tasks. Results demonstrate that JSDRV improves the capabilities of LLMs in the joint tasks, not only outperforming state-of-the-art methods but also generalizing to non-LLMs accommodated as task models.
\end{abstract}

\section{Introduction}

Social media has transformed the ways people access information by facilitating rapid information sharing. However, it has become a fertile ground for nurturing rumors and misinformation due to its lack of systematic moderation~\cite{vosoughi2018spread, cheng2021causal}. Their rampant spread has become a considerable global societal issue that can profoundly impact people's beliefs and normal life~\cite{roozenbeek2019fake}. 

\begin{figure}[t!]
  \centering
  \includegraphics[width=\linewidth]{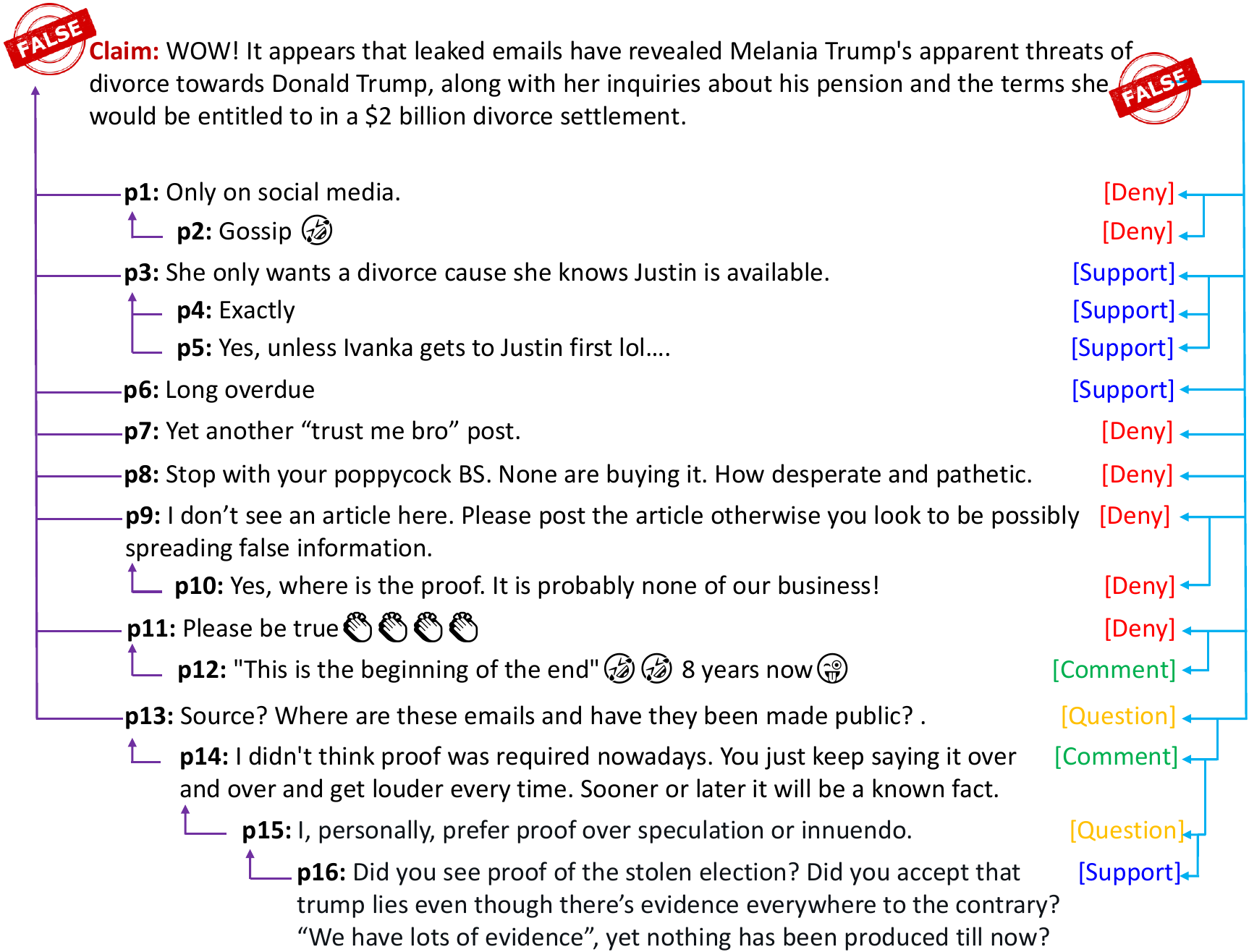}
  \caption{A false rumor claim and the stances of responding posts. The arrow lines indicate the direction of inference for claim veracity and posts stances.}
  \label{fig:intro}
\end{figure}

In general, stance provides insights into the attitudes, opinions, and beliefs of individuals regarding a specific target~\cite{10.1145/3369026,ALDAYEL2021102597}. Stance and rumor are deeply coupled since the stance expressed by social media posts toward a rumorous event, i.e., rumor stance, offers valuable cues for assessing the overall credibility of the target claim. Figure~\ref{fig:intro} shows that the rumor stance helps in understanding the context and the way the rumor is perceived by different users in an intuitive and explainable manner. 

Early research has found that detecting stances expressed in related conversation threads on social media is beneficial for rumor detection and monitoring~\cite{qazvinian-etal-2011-rumor,zhao2015enquiring,10.1371/journal.pone.0150989}. Nowadays, researchers are increasingly focusing on leveraging related stances for improving rumor detection and verification effectiveness~\cite{dungs-etal-2018-rumour,ma2020attention,li-etal-2020-exploiting,10.1007/978-3-030-45439-5_38,yuan2021srlf} as well as performing stance-rumor joint detection tasks through multi-task learning~\cite{ma2018detect,kochkina2018all,wei2019modeling,li-etal-2019-rumor-detection,yu-etal-2020-coupled,yang2022weakly}. 

However, most existing approaches require extensive post-level stance labels and claim-level veracity labels for training stance detection and rumor verification models, respectively, which are very expensive to obtain. While some ``for free'' unsupervised~\cite{kobbe2020unsupervised,allaway2020zero,pick2022stem,li2023tts} methods were developed, they are only for the stance detection task and suffer from poor generalizability due to reliance on crafted features or specific models pre-trained on the data (e.g., online debate) unmatched with rumor stances from social media. More recently, a weakly supervised method~\cite{yang2022weakly} based on multiple instance learning~\cite{DIETTERICH199731,FouldsFrank2010} is proposed to predict the stance of an individual post and the rumor veracity of the claim only being supervised with claim veracity labels. However, determining post stances via dispersing the veracity of a source claim down to stances of many individual posts, as illustrated by Figure~\ref{fig:intro}, is intuitively much more challenging than inferring a source claim's veracity via aggregating post stances in an opposite direction. 

The emergence of disruptive technologies for Large Language Models (LLMs) such as ChatGPT~\cite{NEURIPS2020_1457c0d6} and smaller variants such as Llama~\cite{touvron2023Llama} and ALPACA~\cite{taori2023alpaca} have shown performance and explainability on a par with human or even better in various NLP tasks. 
However, their abilities in rumor-stance related tasks within social media context are still understudied, especially for the joint prediction for detecting posts stances and debunking rumors from claims at the same time. On one hand, LLMs might suit such tasks thanks to their rich pre-trained knowledge and strong zero- and few-shot capabilities; on the other hand, their predictive power based on a large number of rumor-related social media posts might be compromised by noise, unreliability, and lack of supervision. 

To study and unleash the potential of LLMs for this joint task, 
we propose a reinforcement tuning framework for \underline{j}oint \underline{s}tance \underline{d}etection and \underline{r}umor \underline{v}erification (\textbf{JSDRV}) based on LLMs\footnote{Code is released at \url{https://anonymous.4open.science/r/JSDRV-F3CE/}}. The framework contains three complementary parts: the LLM stance detection (SD) network, the reinforcement label selector, and the LLM rumor verification (RV) network. Specifically, assuming merely a small set of seeding veracity labels at claim level, the reinforcement selector chooses high-quality examples for fine-tuning the SD and RV LLMs based on their generated labels and explanations. 
We present an end-to-end joint optimization mechanism to boost the integrated framework.  
Our contributions are summarized as follows:
\begin{itemize}
\item 
We propose a novel LLM-based reinforcement tuning framework to detect stance and verify rumor veracity jointly starting off with a small set of seed instances labeled by humans.
\item 
Our JSDRV framework is generic, which can not only accommodate open or closed LLMs, but also non-LLM-based models as stance detection and rumor verification networks.

\item Extensive experiments on multiple benchmark rumor stance datasets demonstrate that JSDRV outperforms a range of strong baselines, including pretrained language models and fully supervised models on both tasks.
\end{itemize}

\section{Related Work}

\textbf{Rumor Verification.} 
Early studies on rumor verification train supervised classifiers by utilizing content~\cite{yang2012automatic,liu2015real} or contextual features~\cite{zhao2015enquiring,zubiaga2017towards} extracted from claim and related posts from social media. Nowadays, most methods predominately focus on utilizing neural networks such as RNN~\cite{ma2016detecting}, CNNs~\cite{yu2017convolutional}, tree-/graph-based~\cite{lu2020gcan,ma2020attention,rosenfeld2020kernel,lin2021rumor}, transformer-based~\cite{khoo2020interpretable,ma2020debunking,yu-etal-2020-coupled} and adversarial contrastive~\cite{ma2021improving,lin2022detect} models. Researchers also find that stances towards the specific claim shared among social media users, can assist rumor verification by revealing crucial cues and dissemination patterns~\cite{qazvinian-etal-2011-rumor,zhao2015enquiring,10.1371/journal.pone.0150989}, leading to more recent stance-aware approaches for rumor verification~\cite{dungs-etal-2018-rumour,ma2020attention,li-etal-2020-exploiting,10.1007/978-3-030-45439-5_38,yuan2021srlf}. 

Recently, research has leveraged pre-trained language models (PLMs) including LLMs for misinformation-related tasks such as fact checking~\cite{lee-etal-2021-towards,pan-etal-2023-fact,zeng-gao-2023-prompt,zhang-gao:2023:ijcnlp,zhang-gao-2024-reinforcement-retrieval}. 
\citet{lin2023zero} proposed zero-shot prompt learning for rumor detection using a multilingual PLM addressing diverse languages and domains on social media.
Little has been done on using LLMs for rumor stance detection and verification in social media contexts. 


\textbf{Stance Detection.} Stance detection, 
initially relies on hand-crafted features~\cite{lukasik2016hawkes,zubiaga2018discourse}, and later has advanced to use deep learning~\cite{augenstein2016stance,zhang2019stances,liang2021target}. Subsequent research has explored incorporating propagation structure~\cite{zubiaga2016stance,kochkina2017turing}. Recent approaches delve into reinforcement learning~\cite{wei2019topic}, contrastive learning~\cite{liang2022jointcl} and teacher-student models~\cite{li2023tts}. Yet these methods require large annotated corpora for training. To address this limitation, unsupervised~\cite{kobbe2020unsupervised,allaway2020zero,pick2022stem,DBLP:conf/aaai/RanJ23} and weakly supervised model~\cite{yang2022weakly} have emerged, albeit with concerns about their weak detection efficacy and generalizability. 
PLMs elevate stance detection by utilizing variants of BERT~\cite{devlin-etal-2019-bert}, setting new standards this task~\cite{allaway-mckeown-2020-zero,li-etal-2021-p}. ChatGPT demonstrates its accuracy in stance detection~\cite{aiyappa-etal-2023-trust} via zero-shot and few-shot prompt engineering.

\textbf{Rumor-Stance Dual Task.} The rumor-stance dual task commenced from RumorEval shared task series~\cite{derczynski-etal-2017-semeval,gorrell-etal-2019-semeval} as a two-step pipeline, where rumor verification (subtask B) performs veracity prediction based on the claim and SDQC stances of the posts classified in stance detection (subtask A). Then, joint detection has been studied through multi-task learning~\cite{ma2018detect,kochkina2018all,wei2019modeling,li-etal-2019-rumor-detection,yu-etal-2020-coupled}. However, such approach is fully supervised by large training sets labeled for claim veracity and posts stance. \citet{yang2022weakly} proposed a weakly supervised neural model with multiple instance learning only using a full set of veracity-labeled claims for training. We assume only a small set of training examples at claim level as seeds for LLM-based annotation, and learn a policy to select high-quality annotations for fine-tuning SD and RV LLMs.


\section{Problem Statement} \label{sec:ps}

We define a rumor dataset as $\mathcal{C}=\{(c_i,X_i)\}_{i=1}^{|\mathcal{C}|}$, where each instance $(c_i,X_i)$ is a tuple consisting of a source claim $c_i$ and a conversation thread of posts responding to $c_i$ denoted as $X_i=\{x_{i,1}, x_{i,2}, \cdots, x_{i,T}\}$. The posts are presented in a \textit{chronological} order while reply structure may exist via `@user' symbol in the text. We define the dual tasks as follows:
\begin{itemize}
\item \textbf{Stance Detection:} The task is to determine the stance $y_{i,j}$ for each post $x_{i,j} \in X_i$ under claim $c_i$. That is, $f: x_{i,1}x_{i,2} \cdots x_{i,T} \to y_{i,1} y_{i,2} \cdots y_{i,T}$, where $y_{i,j}$ takes one of the Support (S), Deny (D), Question (Q) or Comment (C) stance labels. 
\item \textbf{Rumor Verification:} The task is to classify each claim $c_i$ together with the responding posts into one of the four veracity classes $Y_i$: Non-Rumor (N), True Rumor (T), False Rumor (F), or Unverified Rumor (U). That is, $g: (c_i,X_i) \to Y_i$. 
\end{itemize}

Traditionally, the ground-truth of $y_{i,1} y_{i,2} \cdots y_{i,T}$ and $Y_i$ of all training instances are assumed available for full supervision~\cite{ma2018detect,kochkina2018all,wei2019modeling,li-etal-2019-rumor-detection,yu-etal-2020-coupled}, or only $Y_i$ of each training instance is available for weak supervision~\cite{yang2022weakly}. In contrast, we target a more challenging setting, where only a small set of seeding claims $\mathcal{C}'\in \mathcal{C}$ ($|\mathcal{C}'|\ll|\mathcal{C}|$) are provided with veracity labels, while no post stance is provided for training. 

\section{Our JSDRV Framework}

\begin{figure*}[t!]
 \centering
 \includegraphics[width=0.8\linewidth]{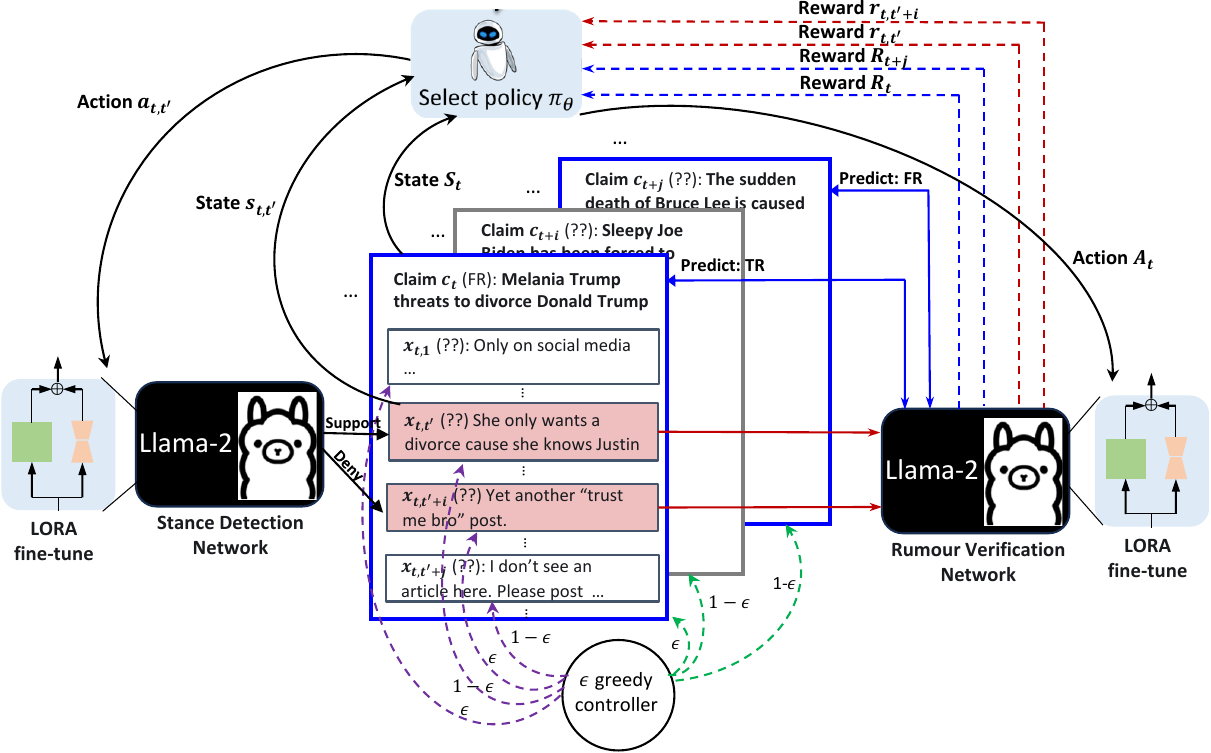}
  \caption{Our reinforcement fine-tuning framework for joint stance detection and rumor verification.}  
  \label{fig:model}
\end{figure*}

Using zero-/few-shot prompting~\cite{radford2019language,NEURIPS2020_1457c0d6} and parameter-efficient fine-tuning (PEFT)~\cite{hu2021lora,pmlr-v97-houlsby19a,lester-etal-2021-power} techniques, LLMs can perform on a par with or even better than traditional supervised models. 
However, using LLMs for the stance-rumor joint tasks faces two major issues: 1) Human labels, especially stance labels of posts in social conversation about a claim, are difficult to obtain. Even if they were abundant, it could be hard to fully utilize them for fine-tuning due to the limit of computing resources; 2) While one could use LLMs to annotate data in scale, LLM-provided labels may be unreliable, subject to further refinement for quality. Thus, how to label and select high-quality instances becomes paramount for effective prompting or fine-tuning.

We propose a reinforcement tuning framework to perform data annotation, selection, and model fine-tuning for the dual tasks, based on a small set of seeding claims with veracity labeled manually. By iteratively refining the selection policy with LLM's annotations, the model prioritizes instances that align with the desired task objectives. With heterogeneous reward functions based on limited ground-truth data, JSDRV enhances the instance selection process, ultimately boosting the overall performance of both tasks. 

Figure~\ref{fig:model} illustrates our end-to-end framework, encompassing an SD network, a reinforcement selection policy model, and an RV network.
We adopt Llama-2 7B version\footnote{While we opt to host local, open-sourced LLMs, our framework is general and can accommodate closed LLMs like GPT-3.5/4 from OpenAI.}~\cite{touvron2023Llama}, an open-source LLM for both SD and RV networks, which can be fine-tuned with PEFT techniques such as LoRA~\cite{hu2021lora}.

\subsection{LLM Stance Detection Network} \label{sec:StanceDetection}
The purpose of the LLM-based SD network is to provide a stance label to any post in ${X_t}$ of a given claim $c_t$. Since the size of $X_t$ can be very large, it is neither necessary nor feasible to use all the posts due to the potential high costs of interaction with LLMs or fine-tuning. We use an $\epsilon$-greedy strategy to pre-select a subset of posts for LLM to annotate before learning our selector policy model ($\S$~\ref{sect:selector}), aiming to facilitate the continuous iterative optimization of the stance LLM.

\paragraph{$\epsilon$-greedy Controller for Post Pre-Selection}
We propose an $\epsilon$ greedy controller to judiciously regulate the quantity of posts submitted for LLM prediction, thereby balancing the exploitation, i.e., choosing posts from $X_t$ by following the time order of posts, and exploration, i.e., selecting posts in $X_t$ randomly. The intuition is that the model strikes to balance the earliness and time span for getting useful posts. Thus, a post $\tilde{x}_{t,t'}$ is sampled at any step $t'$ based on the following trade-off:
\begin{equation}\label{equ:StanceGreedyController}
  \tilde{x}_{t,t'} \sim
\begin{cases}
\text{Next}(X_t) & \text{with prob } \epsilon \\
\text{Uniform}(X_t) & \text{with prob 1- } \epsilon
\end{cases}  
\end{equation}
where $X_t$ is a temporally ordered list of posts as aforementioned, $\text{Next}(.)$ is a function of choosing the next post in the list, and $\text{Uniform}(.)$ is a uniform distribution over all posts in $X_t$. At each step $t'$, we sample a post without replacement from one of the two distributions and feed it into SD LLM for prompting and label generation.

\paragraph{Stance Prompt Learning}

We design a prompt as the instruction to make the SD LLM generate a stance label and a brief explanation for each sampled post, following a required format. 
An example prompt is shown in Figure~\ref{fig:StancePrompt} in Appendix~\ref{sect:SDprompt}. 

\paragraph{Pretraining} \label{sec:StancePretrain} For a better initial quality of labeling, we pre-train the SD LLM with P-stance dataset\footnote{P-stance is a general stance dataset with 3-category stances. We find it still helps SD LLM in 4-class stance classification in our task.}~\cite{li2021p}. The objective is to minimize the negative conditional language modeling for generating correct stance labels.

\subsection{LLM Rumor Verification Network} \label{sec:RumorPrediction}
Posts after pre-selection and annotation by SD network will be further selected by the policy model (\S~\ref{sect:selector}) for determining whether they should be retained or discarded, considering how useful they are to claim veracity prediction by the RV network. 
The RV network also needs to augment labeled instances at claim level, for which a pre-selection of claims is conducted using $\epsilon$-greedy method. Since we have some human labeled claims, the sampling of a claim $\tilde{c}_{t}$ at step $t$ strives to balance exploiting human labeled claims and exploring unlabeled claims both uniformly, which yields:
\begin{equation}\label{equ:RumorGreedyController}
  \tilde{c}_{t} \sim
\begin{cases}
\text{Uniform}(\mathcal{C}') & \text{with prob } \epsilon\\
\text{Uniform}(\mathcal{C}-\mathcal{C}') & \text{with prob 1- } \epsilon
\end{cases}  
\end{equation}
where $\mathcal{C'}$ denotes the labeled claim set.

Given each pre-selected claim and its related posts that are retained by the selection policy model, RV LLM network is then prompted to generate veracity label for the claim and a brief explanation of the decision, considering the claim and posts content and stance. 
An example prompt is shown in Appendix~\ref{sect:RVprompt}. 


\paragraph{Pretraining} \label{sec:RumorPretrain} Similarly, we also pretrain the RV LLM for better initial quality using the small manually labeled claim set $\mathcal{C'}$. Without post stance labels, we just feed posts content with the claim into the RV LLM, which is trained to minimize the negative likelihood of predicted labels and ground truth.

\subsection{Selector Policy}\label{sect:selector}
We design a selector policy to transform input states, i.e., annotated instances that are pre-selected, to their corresponding actions, i.e., decisions to discard or retain an instance. 
For optimizing the selector's policy $\pi_\theta$ with parameter $\theta$, each step corresponds to sampling an annotated claim followed by a sequence of sub-steps sampling annotated posts, and receiving rewards from the RV network. 

We formulate a two-level Markov Decision Process (MDP) with the following elements: (1) $\{S_t\}$ and $\{s_{t,t'}\}$ correspond to a sequence of states at claim and post level, respectively; (2) $\{A_t\}$ and $\{a_{t,t'}\}$ correspond to a sequence of actions for sampled claims and posts, respectively, for deciding whether to keep the current instance; (3) $\{R_t\}$ and $\{r_{t,t'}\}$ are rewards received after taking an action for the respective level. This reward incorporates predictions based on instances with and without human labels.
For easing presentation, we will use a unified notation to describe these elements at both levels, that is, $\varsigma$, $\alpha$, $\gamma$, and $\tau$ denote state, action, reward, and time step, respectively, but note that in practice each of them is separated into two versions corresponding to claim and post as mentioned. 

\paragraph{State} 
State $\varsigma_{\tau}$ denotes the current status at step $\tau$ after the previous instances, i.e., claims or posts, sampled up to $\tau-1$. It contains the representations of three parts: the current claim $\tilde{c}_{\tau}$, the selected instances thus far, and the prediction explanation for the current instance. This yields $\varsigma_{\tau} = \left[\tilde{\mathbf{c}}_{\tau}, \mathbf{C}_{\tau-1}, \mathbf{E}_{\tau}\right]$,
where $\tilde{\mathbf{c}}_{\tau}$, $\mathbf{C}_{\tau-1}$ and $\mathbf{E}_{\tau}$ denote the embedding of $\tilde{c}_{\tau}$, the context by averaging the embeddings of selected instances up to $\tau-1$, and the embedding of explanation for the current instance, respectively. The embeddings are obtained through RoBERTa~\cite{liu2019roberta}.

\paragraph{Action} 
An action $\alpha_{\tau}$ is sampled from $\pi_\theta$ stochastically given the state $\varsigma_{\tau}$. Specifically, the policy network will output a probability distribution over action space \{\textit{discard}, \textit{retain}\}, which yields:
\begin{equation}\label{equ:policy}
    \pi_\theta(\alpha_{\tau}, \varsigma_{\tau}) = \sigma(w_2 \cdot \text{ReLU}(w_1\cdot \varsigma_{\tau}))
\end{equation}
where $\theta=\{w_1,w_2\}$ are the weights of policy network and $\sigma$ is the sigmoid activation function. Then, the action $\alpha_{\tau}$ is sampled according to the output probability: $\alpha_{\tau}\sim\pi_\theta(\alpha_{\tau}, \varsigma_{\tau})$. 


\paragraph{Reward} 
For a selected claim with human label, the reward considers congruence between the RV network prediction and its ground truth, since the contribution of each selected post under the claim can be reflected by claim veracity prediction; For a selected claim without ground truth, the reward considers how well the stance distribution of its sampled posts conforms to the posts stance distribution of those claims in $\mathcal{C'}$ that have the same veracity label as the predicted label of the selected claim. This yields:
\begin{equation}\label{equ:Reward}
\small
  \gamma_{\tau} =
\begin{cases}
 \mathbb{E}(\cos(\hat{\mathbf{Y}}_{\tilde{c}_{\tau}}, \mathbf{Y}_{\tilde{c}_{\tau}})) & \text{if }{\tilde{c}_{\tau}\in \mathcal{C'}} \\
\mathbb{E}(\cos(\overline{\hat{\mathbf{y}}_{\tilde{c}_{\tau},\tilde{x}_{\tau,*}}}, \overline{\hat{\mathbf{y}}_{c\in\mathcal{C'},x_{c,*}}}|\hat{Y}_{\tilde{c}_{\tau}}=Y_c)) & \text{otherwise}
\end{cases}  
\end{equation}
Here, $\hat{\mathbf{Y}}_{\tilde{c}_{\tau}}$ and $\mathbf{Y}_{\tilde{c}_{\tau}}$ are respectively the predicted and ground-truth veracity distributions of sampled claim $\tilde{c}_{\tau}$; $\overline{\hat{\mathbf{y}}_{\tilde{c}_{\tau},\tilde{x}_{\tau,*}}}$ is the mean of stance distributions predicted on the selected posts $\tilde{x}_{\tau,*}$ under $\tilde{c}_{\tau}$; $\overline{\hat{\mathbf{y}}_{c\in\mathcal{C'},x_{c,*}}}$ is the mean of stance distributions predicted on all the posts $x_{c,*}$ over all the claims $c \in \mathcal{C'}$, of which the veracity label is same as the predicted veracity of $\tilde{c}_{\tau}$;
and $\mathbb{E}(.)$ is a sign function that turns the cosine similarity of two distributions to -1, 0 or 1 depending on the sign of similarity. 
The reward encourages the model to retain the instances (i.e., claims and posts) that can help veracity prediction keep close to the human-labeled claims providing similar stance distribution.

\subsection{Model Training}
Our training is an end-to-end joint optimization process, which involves alternating training on policy network and LLM-based SD and RV networks in each epoch.

\paragraph{Policy Network} We employ the widely used offline optimization method~\cite{sutton2018reinforcement} to maximize expected accumulative reward $\mathcal{R}_t$:
\begin{dmath}\label{equ:JointOptRL}
    \mathcal{R}_{t} = \frac{1}{t} \sum_{i=1}^{t} \left(R_i \log(\pi_\theta(A_i,S_i)) + \frac{1}{t'}\sum_{j=1}^{t'} r_{i,j}\log(\pi_\theta(a_{i,j},s_{i,j}))\right)
\end{dmath}
where $R_i$ and $r_{i,j}$ are calculated by Equation~\ref{equ:Reward}, and $t$ and $t'$ respectively denote the current time step at claim and post levels. The policy network is updated after a claim and its posts are selected.

\paragraph{LLMs} 
We fine-tune the LLM-based SD and RV networks using the annotations (human or machine labeled) of selected instances by minimizing standard Negative Log Likelihood loss in language model training~\cite{kanamori2010deformation}. In each epoch, the two LLMs are fine-tuned only once following the selection process. Note that we can skip fine-tuning them but use the selected instances for few-shot in-context learning.


\paragraph{Training Procedure} The training detail is depicted as Algorithm~\ref{alg:opt}. JSDRV is trained with an end-to-end fashion by optimizing the two-level selector policy model for claim selection and post selection and the SD and RV LLMs.

\begin{algorithm*}
  \caption{The JSDRV Model Training}\label{alg:opt}
  \begin{algorithmic}[1]
    \State \textbf{Input:} A rumor dataset $\mathcal{C}$, seeding claims $\mathcal{C'}\in \mathcal{C}$, P-stance dataset, SD network, RV network, $\epsilon$, $\pi_\theta$ policy network.
\State Pretrain SD network with P-stance dataset according to $\S$~\ref{sec:StancePretrain}, and RV network with seeding claims according to $\S$~\ref{sec:RumorPretrain}.
  \For{$t \gets 1$ to $|\mathcal{C}|$}
    \If {termination condition meets} 
        \State break. 
    \EndIf
    \State Sample a claim $\Tilde{c}_t$ according to Equation~\ref{equ:RumorGreedyController}.
    \For{$t' \gets 1$ to $|X_{\tilde{c}_t}|$}
    \If {termination condition meets}
        \State break
    \EndIf
    \State Sample a post according to Equation~\ref{equ:StanceGreedyController}.
    \State Predict post stance according to $\S$~\ref{sec:StanceDetection}.
    \State Calculate post-level reward according to Equation~\ref{equ:Reward}.
    \EndFor
    \State Predict claim veracity according to $\S$~\ref{sec:RumorPrediction}.
    \State Calculate claim-level reward according to Equation~\ref{equ:Reward}.
    \State Update parameters of $\pi_\theta$ based on Equation~\ref{equ:JointOptRL}.
    \State Fine-tune SD network (w/ LoRA). 
    \State Fine-tune RV network (w/ LoRA). 
  \EndFor
  \State \textbf{return} SD network, $\pi_\theta$, RV network.
  \end{algorithmic}
\end{algorithm*}

\paragraph{Termination condition} The $\epsilon$ greedy process can terminate automatically. For the termination condition, we utilize the reward function described in Equation~\ref{equ:Reward}. The selection process stops if the model receives reward $\gamma_{\tau}=1$ for $N$ continuous steps, for which we set $N$ as 100 tuned on the validation set.

\section{Experiments and Results}

\subsection{Datasets and Setup}
\paragraph{Datasets} For model training, we utilize three public rumor verification benchmark datasets, Twitter15/16 (T15/16)~\cite{ma2017detect} and PHEME (PH)\footnote{\url{https://figshare.com/articles/PHEME_dataset_of_rumours_and_non-rumours/4010619}.}, where only claim veracity labels are provided. 
Since both stance label and rumor veracity are required for testing, we resort to RumorEval-S\footnote{\url{https://github.com/2302Jerry/Data-Repo}}~\cite{yang2022weakly} and SemEval-8~\cite{derczynski-etal-2017-semeval} datasets with labeled claim veracity and posts stance. Details of the datasets are provided in Table~\ref{tab:TrainData} and Table~\ref{tab:TestData} in Appendix~\ref{sect:datastat}.

\paragraph{Setup}
We hold out 20\% instances of test sets as validation sets (Val), for training models that require stance labels and tuning hyper-parameters. We use micro-averaged, macro-averaged F1 score, and class-specific F-measure as evaluation metrics considering imbalanced class distributions~\cite{zubiaga2016stance}. We use public codes for baselines or re-implement them if codes are not released. 
We use 50\% claims in the training set as seeding claims for JSDRV for the main results and examine how the percentage affects its performance (see Appendix~\ref{sect:seed}). We set $\epsilon=0.3$ with validation data. The detail of other hyper-parameters setup is provided in Appendix~\ref{app:LlamaArgs}.

\subsection{Stance Detection Performance}

Since our SD network does not need post stance label, we choose the following unsupervised, supervised, and weakly supervised baselines, which are detailed in Appendix~\ref{app:stanceBase}:
(1) \textbf{TGA}~\cite{allaway2020zero}; 
(2) \textbf{BerTweet}~\cite{nguyen2020bertweet}; 
(3) \textbf{Llama 2-ST}~\cite{touvron2023Llama};
(4) \textbf{Llama 2-MT}~\cite{touvron2023Llama};
(5) \textbf{BiGRU}~\cite{augenstein2016stance};
(6) \textbf{BrLSTM}~\cite{kochkina2017turing}; 
(7) \textbf{MT-GRU}~\cite{ma2018detect};
(8) \textbf{JointCL}~\cite{liang2022jointcl};
(9) \textbf{SRLF}~\cite{yuan2021srlf}; 
(10) \textbf{TD-MIL}~\cite{yang2022weakly}.
We use \textbf{model (\textsc{DataSet})} to denote \textbf{model} trained on \textsc{DataSet}\footnote{We also train JSDRV on the validation set for fairly compared with supervised methods that need stance labels.}.

For the stance detection baselines, we use the original source code of TGA. The first group refers to unsupervised baselines, while BiGRU, BrLSTM, MT-GRU, and JointCL in the second group are four popular supervised stance detection baselines. SRLF and TD-MIL are weakly supervised models that do not need stance annotation. We use the validation dataset to train the JSDRV (Val) variant for fair comparison with the supervised models. This is because there is no stance annotations in the training set.

\begin{table*}[t!]
\setlength{\abovecaptionskip}{0.1cm}
  \centering
  \resizebox{0.98\textwidth}{!}{
    \begin{tabular}{l|cc|cccc|cc|cccc}
    \toprule
    Dataset & \multicolumn{6}{c|}{RumorEval-S}  & \multicolumn{6}{c}{SemEval-8} \\
    \midrule
    \multirow{2}[4]{*}{Method} & \multirow{2}[4]{*}{MicF} & \multirow{2}[4]{*}{MacF} & S     & D     & Q     & C     & \multirow{2}[4]{*}{MicF} & \multirow{2}[4]{*}{MacF} & S     & D     & Q     & C \\
\cmidrule{4-7}\cmidrule{10-13}       &       &       & F1    & F1    & F1    & F1       &       &       & F1    & F1    & F1    & F1 \\
    \midrule
    TGA & -       & 0.324  & 0.301  & 0.168  & 0.342  & 0.486     & 0.383  & 0.344  & 0.278  & 0.162  & 0.480  & 0.456  \\
    BerTweet  & 0.619  & 0.492  & 0.497  & 0.203  & 0.513  & 0.753  & 0.611  & 0.428  & 0.512  & 0.131  & 0.326  & 0.742  \\
    Llama 2-ST  & 0.630  & 0.500  & 0.501  & 0.203  & 0.532  & 0.763   & 0.631  & 0.471  & 0.533  & 0.138  & 0.472  & 0.740  \\
    Llama 2-MT  & 0.632  & 0.500  & 0.502  & 0.199  & 0.533  & 0.766   & 0.630  & 0.473  & 0.534  & 0.142  & 0.471  & 0.742 \\
    \midrule
    BiGRU (Val)  & 0.630  & 0.417  & 0.392  & 0.162  & 0.360  & 0.754    & 0.633  & 0.416  & 0.460  & 0.168  & 0.328  & 0.708  \\
    BrLSTM (Val)   & 0.660  & 0.420  & 0.460  & 0.000  & 0.391  & 0.758   & 0.665  & 0.401  & 0.493  & 0.000  & 0.381  & 0.730  \\
    MT-GRU (Val)   & 0.636  & 0.432  & 0.313  & 0.156  & 0.506  & 0.748  & 0.630  & 0.413  & 0.498  & 0.116  & 0.312  & 0.729  \\
    JointCL (Val)   & 0.639  & 0.505  & 0.532  & 0.210  & 0.516  & 0.760  & 0.640  & 0.475  & 0.536  & 0.136  & 0.478  & 0.751  \\
    \midrule
    SRLF (PH)   & 0.606  & 0.479  & 0.492  & 0.280  & 0.468  & 0.676  & 0.510  & 0.393  & 0.328  & 0.205  & 0.420  & 0.619  \\
    TD-MIL (PH)  & 0.691 & 0.434 & 0.344 & 0.179 & 0.467 & 0.767 & 0.651 & 0.426 & 0.335 & 0.175 & 0.430 & 0.763 \\
    \midrule
     \textbf{JSDRV-Bert (Val)}  & 0.668  & 0.541  & 0.531  & 0.316  & 0.562  & 0.755   & 0.658  & 0.489  & 0.540  & 0.173  & 0.482  & 0.761  \\
    \textbf{JSDRV-Bert (T15)}   & 0.680  & 0.562  & 0.534  & 0.380  & 0.570  & 0.764   & 0.671  & 0.498  & 0.549  & 0.169  & 0.490  & 0.784  \\
    \textbf{JSDRV-Bert (T16)}  & 0.681  & 0.560  & 0.527  & 0.381  & 0.573  & 0.759  & 0.673  & 0.500  & 0.550  & 0.170  & 0.490  & 0.790  \\
    \textbf{JSDRV-Bert (PH)}  & 0.683  & 0.565  & 0.535  & 0.383  & 0.573  & 0.765   & 0.780  & 0.502  & 0.556  & 0.170  & 0.493  & 0.789  \\
    \midrule
    \textbf{JSDRV (Val)}  & 0.672  & 0.550  & 0.536  & 0.310  & 0.576  & 0.779   & 0.673  & 0.496  & 0.542  & 0.168  & 0.483  & 0.790  \\    
    \textbf{JSDRV (T15)}   & 0.696  & 0.576  & 0.535  & 0.383  & 0.586  & \textbf{0.801} & 0.693  & 0.506  & 0.558  & 0.170  & 0.496  & 0.798  \\
    \textbf{JSDRV (T16)}  & 0.697  & 0.574  & 0.536  & 0.380  & 0.580  & 0.798   & 0.696  & 0.507  & 0.560  & 0.173  & 0.498  & 0.796  \\
    \textbf{JSDRV (PH)} & \textbf{0.723} & \textbf{0.605} & \textbf{0.546} & \textbf{0.476} & \textbf{0.595} & \textbf{0.801} & \textbf{0.705} & \textbf{0.522} & \textbf{0.563} & \textbf{0.216} & \textbf{0.506} & \textbf{0.801} \\
    \bottomrule
    \end{tabular}%
    }
  \caption{Stance detection results. JSDRV models use 50\% claims in training sets as seeding claims.}
  \label{tab:StanceResult}%
\end{table*}%


From Table~\ref{tab:StanceResult}, we observe that: 1) In zero-shot models, TGA performs worst as it is pretrained on specific topics and cannot generalize well to Twitter data; 
BerTweet, which is fine-tuned on enormous Twitter datasets, outperforms TGA; 
Llama 2-ST and -MT outperform BerTweet, indicating that the LLM has promising zero-shot capability to detect stances. 2) In fully supervised baselines, Bi-GRU based on a sequential architecture performs worst; BrLSTM, benefiting from propagation structure, makes improvements, but is unable to detect deny stance as it is sensitive to such infrequent class in the training data; JointCL leveraging both context- and target-aware features outperforms MT-GRU which is sequential. 3) While both TD-MIL and SRL are weakly supervised only using claim labels, TD-MIL benefits from propagation information while SRLF cannot. 4) Trained on the validation set, JSDRV (Val) and its BERT-based variant are comparable to TD-MIL trained on the full training set, indicating JSDRV is less demanding on labeled data.

JSDRV outperforms all the baselines on the corresponding datasets. When LLM is replaced by BERT in JSDRV, there is a performance drop but JSDRV-Bert is still superior to baselines, suggesting it can be generalized to non-LLM task models.
JSDRV (PH) performs the best, beating its counterparts trained on T15/16 datasets with large margins due to the much larger size of PH dataset.

\subsection{Rumor Verification Performance}
We compare to unsupervised, supervised, weakly supervised, and multi-task rumor verification baselines:
(1) \textbf{BerTweet}~\cite{nguyen2020bertweet};
(2) \textbf{Llama 2-ST}~\cite{touvron2023Llama};
(3) \textbf{Llama 2-MT}~\cite{touvron2023Llama};
(4) \textbf{GCAN}~\cite{lu2020gcan}; 
(5) \textbf{TD-RvNN}~\cite{ma2020attention};
(6) \textbf{PLAN}~\cite{khoo2020interpretable};
(7) \textbf{DDGCN}~\cite{sun2022ddgcn};
(8) \textbf{SRLF}~\cite{yuan2021srlf};
(9) \textbf{TD-MIL}~\cite{yang2022weakly};
(10) \textbf{MTL2}~\cite{kochkina2018all};
(11) \textbf{MT-GRU}~\cite{ma2018detect}.
The details are depicted in Appendix~\ref{app:rumorBase}. In Table~\ref{tab:RumorResult}, we report the best rumor verification results obtained across different training and test datasets. MT-GRU and MTL2 require both rumor and stance labels for training. So, we train them on the validation set,  which has both rumor and stance labels. We also compare JSDRV (Val) with MT-GRU and MTL2.

\begin{table*}[t!]
\setlength{\abovecaptionskip}{0.1cm}
  \centering
  \resizebox{0.92\textwidth}{!}{
    \begin{tabular}{l|cc|cccc|cc|ccc}
    \toprule
    Dataset & \multicolumn{6}{c|}{RumorEval-S}                       & \multicolumn{5}{c}{SemEval-8} \\
    \midrule
    \multirow{2}[4]{*}{Method} & \multirow{2}[4]{*}{MicF} & \multirow{2}[4]{*}{MacF} & T     & F     & U     & \multicolumn{1}{c|}{N} & \multirow{2}[4]{*}{MicF} & \multirow{2}[4]{*}{MacF} & T     & F     & U \\
\cmidrule{4-7}\cmidrule{10-12}           &       &       & F1    & F1    & F1    & \multicolumn{1}{c|}{F1}   &       &       & F1    & F1    & F1 \\
    \midrule
    BerTweet  & 0.760  & 0.452  & 0.641  & 0.293  & 0.367  & 0.460  & 0.755  & 0.427  & 0.630  & 0.256  & 0.395  \\
    Llama 2-ST  & 0.754  & 0.450  & 0.660  & 0.271  & 0.400  & 0.469   & 0.746  & 0.424  & 0.632  & 0.260  & 0.380  \\
    Llama 2-MT  & 0.758  & 0.465  & 0.678  & 0.301  & 0.403  & 0.478  & 0.756  & 0.427  & 0.635  & 0.263  & 0.382  \\
    \midrule
    GCAN (PH)  & 0.645  & 0.253  & 0.249  & 0.310  & 0.113  & 0.339  & 0.645  & 0.255  & 0.241  & 0.326  & 0.198  \\
    TD-RvNN (PH)  & 0.753  & 0.677  & 0.755  & 0.666  & 0.673  & 0.615  & 0.748  & 0.694  & 0.712  & 0.617  & 0.753  \\
    PLAN (PH)  & 0.800  & 0.743  & 0.819  & 0.760  & 0.780  & 0.612  & 0.794  & 0.720  & 0.741  & 0.694  & 0.726  \\ 
    DDGCN (PH)  & 0.759  & 0.663  & 0.713  & 0.663  & 0.669  & 0.607  & 0.755  & 0.685  & 0.709  & 0.624  & 0.723  \\
    \midrule
    SRLF (PH)  & 0.742  & 0.447  & 0.667  & 0.290  & 0.381  & 0.452  & 0.742  & 0.423  & 0.635  & 0.249  & 0.386  \\
    TD-MIL (PH) & 0.809 & 0.776 & 0.826 & 0.659 & 0.669 & 0.852 & 0.798 & 0.741 & 0.741 & 0.672 & 0.810\\
    \midrule
    MTL2 (Val)  & 0.653  & 0.430  & 0.622  & 0.279  & 0.352  & 0.457  & 0.651  & 0.433  & 0.640  & 0.289  & 0.372  \\
    MT-GRU (Val) & 0.768  & 0.452  & 0.662  & 0.298  & 0.373  & 0.457   & 0.761  & 0.428  & 0.639  & 0.254  & 0.391  \\
    \midrule
     \textbf{JSDRV-Bert (Val)} & 0.752  & 0.580  & 0.758  & 0.488  & 0.500  & 0.574  & 0.750  & 0.577  & 0.746  & 0.493  & 0.492  \\
    \textbf{JSDRV-Bert (T15)} & 0.803  & 0.770  & 0.800  & 0.750  & 0.753  & 0.777  & 0.783  & 0.730  & 0.758  & 0.701  & 0.731  \\
    \textbf{JSDRV-Bert (T16)}  & 0.810  & 0.776  & 0.802  & 0.751  & 0.751  & 0.800  & 0.785  & 0.732  & 0.760  & 0.710  & 0.726  \\
    \textbf{JSDRV-Bert (PH)}  & 0.813  & 0.780  & 0.810  & 0.762  & 0.756  & 0.792  & 0.796  & 0.746  & 0.779  & 0.712  & 0.747  \\
    \midrule
    \textbf{JSDRV (Val)} & 0.763  & 0.592  & 0.765  & 0.486  & 0.510  & 0.612  & 0.756  & 0.579  & 0.749  & 0.492  & 0.495  \\
    \textbf{JSDRV (T15)} & 0.828  & 0.786  & \textbf{0.830} & 0.759  & 0.788  & 0.762 & 0.829  & 0.755  & 0.769  & 0.731  & 0.766  \\
    \textbf{JSDRV (T16)}  & 0.838  & 0.786  & 0.829  & \textbf{0.770} & 0.782  & 0.765 & 0.830  & 0.768  & 0.800  & 0.734  & 0.770  \\
    \textbf{JSDRV (PH)} & \textbf{0.842 } & \textbf{0.804} & 0.829  & \textbf{0.774} & \textbf{0.824} & \textbf{0.787} & \textbf{0.834} & \textbf{0.784}  & \textbf{0.820} & \textbf{0.741} & \textbf{0.792} \\
    \bottomrule
    \end{tabular}%
  }
  \caption{Rumor verification results. JSDRV models use 50\% claims in training sets as seeding claims.}    
  \label{tab:RumorResult}%
\end{table*}%

From Table~\ref{tab:RumorResult}, we observe a similar trend as the stance detection task. While Llama 2-ST and -MT outperform BerTweet, their performance is still on a par, indicating directly prompting Llama 2 to perform both tasks together is similar to doing them separately. Among supervised baselines, GCAN performs worst because it only considers local structure while structure-aware models such as TD-RvNN, PLAN and DDGCN appear much better.
MTL2 and MT-GRU are multi-task frameworks that require both labels, so we trained them on validation sets. 
JSDRV (Val), trained on the same dataset only using a small set of veracity-labeled claims, outperforms MT-GRU on MacF and class-level F1, indicating that JSDRV is effective in multi-task learning.

Similarly, JSDRV outperforms all the baselines on the corresponding datasets for the RV task, and is still well generalized to the non-LLM-based task models with the LLMs replaced by BERT. JSDRV (PH) outperforms all the baselines regardless of datasets used and gets 3.2\%/17.1\% MicF/MacF improvement over the best baseline TD-MIL (PH) on RumorEval-S. 

\subsection{Ablation Study}

For the ablation, we separate the components in JSDRV. (\textbf{RSS}): reinforced stance selector; (\textbf{RVS}): reinforced veracity selector; (\textbf{FTSD}): fine-tuning SD; (\textbf{FTRV}): fine-tuning RV; (\textbf{PTSD}): pretraining SD; (\textbf{PTRV}): pretraining RV; (\textbf{epSD}): $\epsilon$ greedy control for SD; (\textbf{epRV}): $\epsilon$ greedy control for RV.

Table~\ref{tab:AblationStudy} shows that except removing \textbf{epSD} and \textbf{epRV}, removing other components decreases the performance, indicating they are useful and the pre-selection with $\epsilon$ greedy method can maintain performance with a little cost. Removing \textbf{RSS} and \textbf{RVS} drops the most, meaning that the reinforced selector is the most vial. 
Removing \textbf{FTRV} and \textbf{FTSD} drops more than removing \textbf{PTSD} and \textbf{PTRV} respectively, indicating that joint fine-tuning is more important.

\begin{table}[t!]
  \centering
  \resizebox{0.42\textwidth}{!}{
    \begin{tabular}{l|cc|cc}
    \toprule
    \multirow{2}[4]{*}{Method} & \multicolumn{2}{c|}{Rumor Verification} & \multicolumn{2}{c}{Stance Detection} \\
\cmidrule{2-5}            & MicF  & MacF    & MicF  & MacF \\
    \midrule
    JSDRV  & 0.842 & 0.804  & 0.723 & 0.605 \\
    \midrule
    - RSS     & 0.820  & 0.783   & 0.700  & 0.571  \\
    - RVS    & 0.826  & 0.790   & 0.705  & 0.573  \\
    - FTSD     & 0.828 & 0.790  & 0.709 & 0.578 \\
    - FTRV     & 0.830 & 0.793  & 0.710  & 0.583 \\
    - PTSD      & 0.831  & 0.794   & 0.712  & 0.589  \\
    - PTRV     & 0.833  & 0.795   & 0.714  & 0.600  \\ 
    - epSD     & 0.844 & 0.805  & 0.725  & 0.606 \\
    - epRV    & 0.843 & 0.806  & 0.723  & 0.605 \\        
    \bottomrule
    \end{tabular}%
    }
  \caption{Ablation study on RumorEval-S dataset.}    
  \label{tab:AblationStudy}%
\end{table}%

\subsection{Analysis}

\begin{figure}[t!]
\setlength{\abovecaptionskip}{-0.1cm}
\centering
\subfigure[Original]{
\includegraphics[scale=0.227]{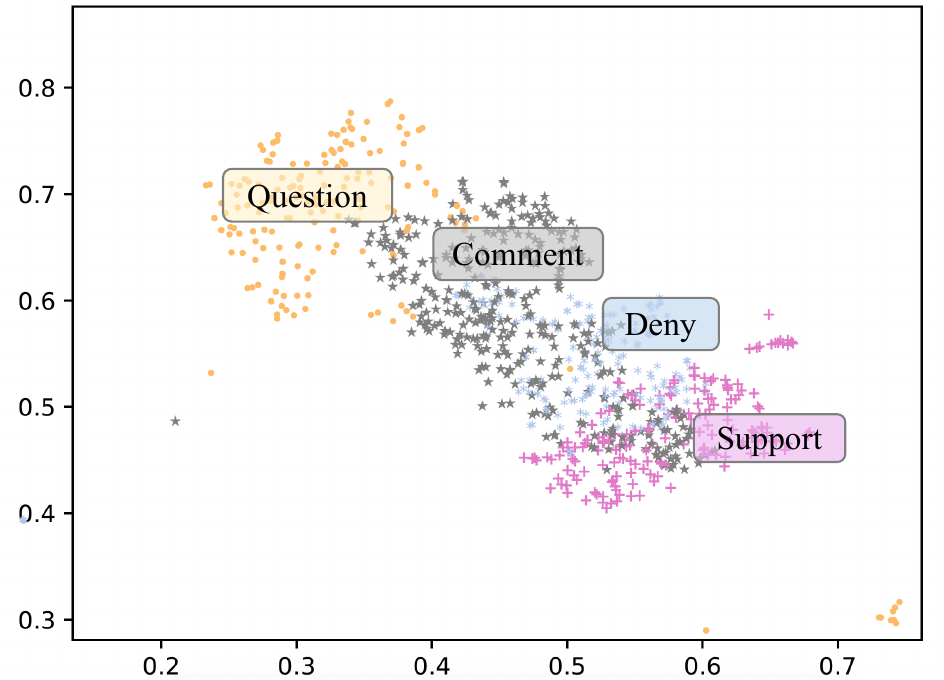}
\label{fig:origianlstance}
}
\subfigure[Selected by JSDRV]{
\includegraphics[scale=0.23]{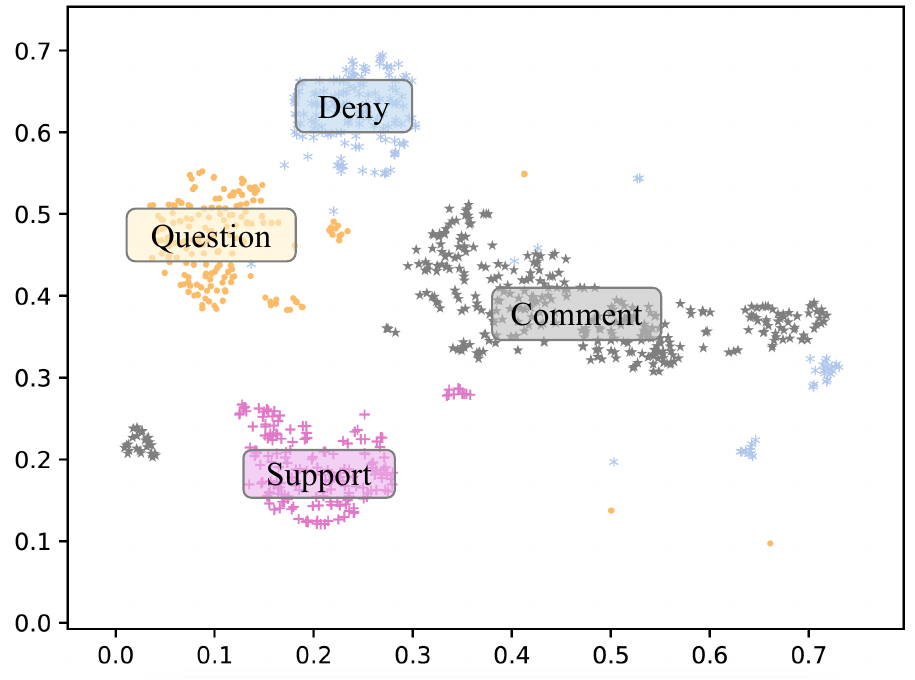}
\label{fig:selectedstance}
}
\caption{Visualization of stance distributions of original and selected posts.} 
\label{fig:VisualizeStance}
\end{figure}

We plot posts with their stances on RumorEval-S dataset with tSNE~\cite{van2008visualizing} to show the effect of post stance selection. We input each post as "[CLS] post [SEP] stance [SEP] reason" into RoBERTa-base and take [CLS] token representations to plot Figure~\ref{fig:VisualizeStance}. We observe that the  selected stances are separated much better, indicating that JSDRV can differentiate stances and may help rumor verification. 

We also plot example outputs of JSDRV in Figure~\ref{fig:CaseStudy} in Appendix~\ref{sect:case} and find that the selector policy can choose posts with high-quality stances annotated by stance LLM, which provides useful cues to rumor verification.
More analyses on the impact of seeding claims and $\epsilon$ are in Appendix~\ref{sect:seed}. 

\vspace{-0.1cm}




\section{Conclusion and Future Work}
 
We introduced a novel reinforcement tuning framework, JSDRV, for joint stance detection and rumor verification tasks based on LLMs. By leveraging LLMs as annotators and employing a reinforcement label selector, we effectively fine-tune the LLMs for both tasks supervised only with minimal human-labeled claims. Our experiments on benchmark datasets demonstrated the superiority of JSDRV over existing methods, suggesting its promising potential for addressing the challenges of joint stance detection and rumor verification in social media environments.
Our future work aims to develop unsupervised methods for both tasks.

\section*{Limitations}



JSDRV relies on the SD and RV networks to provide class probability distributions for calculating rewards. While some LLMs like Llama 2 can return word distribution, converting it to the distribution of a limited number of classes may compromise accuracy. Moreover, LLMs like ChatGPT, which do not return such distributions, could hinder the applicability and practical use of our method.

In JSDRV, frequent interactions with SD and RV networks, particularly when they rely on closed LLM services, can introduce overhead during training. This overhead may impact training efficiency. 

Despite demanding much less labeled data, JSDRV necessitates human-labeled claims, posing challenges in obtaining them, especially in certain domains such as healthcare and science. This constraint may limit its applicability in scenarios where labeled claims are scarce. Therefore, our future work aims to develop unsupervised methods for both tasks.

\section*{Acknowledgements}
This work is partially supported by National Natural Science Foundation of China Young Scientists Fund (No. 62206233) and Hong Kong RGC ECS (No. 22200722).

\section*{Ethical Considerations}
While our research utilizes publicly accessible datasets, using social media conversations for debunking rumors and detecting user stances may raise privacy concerns.  While our approach does not necessitate access to sensitive user data, we anonymized all social media posts, ensuring user information is invisible and unusable by others.   

Both tasks studied in this paper hold social implications. Key considerations include system reliability and the potential for mislabeling information and misleading users. To address this, we will establish responsible policies for code dissemination, aligning with ethical standards.

\bibliography{custom}

\begin{thebibliography}{75}
\expandafter\ifx\csname natexlab\endcsname\relax\def\natexlab#1{#1}\fi

\bibitem[{Aiyappa et~al.(2023)Aiyappa, An, Kwak, and Ahn}]{aiyappa-etal-2023-trust}
Rachith Aiyappa, Jisun An, Haewoon Kwak, and Yong-yeol Ahn. 2023.
\newblock \href {https://doi.org/10.18653/v1/2023.trustnlp-1.5} {Can we trust the evaluation on {C}hat{GPT}?}
\newblock In \emph{Proceedings of the 3rd Workshop on Trustworthy Natural Language Processing (TrustNLP 2023)}, pages 47--54, Toronto, Canada. Association for Computational Linguistics.

\bibitem[{ALDayel and Magdy(2021)}]{ALDAYEL2021102597}
Abeer ALDayel and Walid Magdy. 2021.
\newblock \href {https://doi.org/https://doi.org/10.1016/j.ipm.2021.102597} {Stance detection on social media: State of the art and trends}.
\newblock \emph{Information Processing \& Management}, 58(4):102597.

\bibitem[{Allaway and McKeown(2020{\natexlab{a}})}]{allaway2020zero}
Emily Allaway and Kathleen McKeown. 2020{\natexlab{a}}.
\newblock Zero-shot stance detection: A dataset and model using generalized topic representations.
\newblock In \emph{Proceedings of the 2020 Conference on Empirical Methods in Natural Language Processing (EMNLP)}, pages 8913--8931.

\bibitem[{Allaway and McKeown(2020{\natexlab{b}})}]{allaway-mckeown-2020-zero}
Emily Allaway and Kathleen McKeown. 2020{\natexlab{b}}.
\newblock \href {https://doi.org/10.18653/v1/2020.emnlp-main.717} {{Z}ero-{S}hot {S}tance {D}etection: {A} {D}ataset and {M}odel using {G}eneralized {T}opic {R}epresentations}.
\newblock In \emph{Proceedings of the 2020 Conference on Empirical Methods in Natural Language Processing (EMNLP)}, pages 8913--8931, Online. Association for Computational Linguistics.

\bibitem[{Augenstein et~al.(2016)Augenstein, Rockt{\"a}schel, Vlachos, and Bontcheva}]{augenstein2016stance}
Isabelle Augenstein, Tim Rockt{\"a}schel, Andreas Vlachos, and Kalina Bontcheva. 2016.
\newblock Stance detection with bidirectional conditional encoding.
\newblock In \emph{Proceedings of the 2016 Conference on Empirical Methods in Natural Language Processing}, pages 876--885.

\bibitem[{Brown et~al.(2020)Brown, Mann, Ryder, Subbiah, Kaplan, Dhariwal, Neelakantan, Shyam, Sastry, Askell, Agarwal, Herbert-Voss, Krueger, Henighan, Child, Ramesh, Ziegler, Wu, Winter, Hesse, Chen, Sigler, Litwin, Gray, Chess, Clark, Berner, McCandlish, Radford, Sutskever, and Amodei}]{NEURIPS2020_1457c0d6}
Tom Brown, Benjamin Mann, Nick Ryder, Melanie Subbiah, Jared~D Kaplan, Prafulla Dhariwal, Arvind Neelakantan, Pranav Shyam, Girish Sastry, Amanda Askell, Sandhini Agarwal, Ariel Herbert-Voss, Gretchen Krueger, Tom Henighan, Rewon Child, Aditya Ramesh, Daniel Ziegler, Jeffrey Wu, Clemens Winter, Chris Hesse, Mark Chen, Eric Sigler, Mateusz Litwin, Scott Gray, Benjamin Chess, Jack Clark, Christopher Berner, Sam McCandlish, Alec Radford, Ilya Sutskever, and Dario Amodei. 2020.
\newblock \href {https://proceedings.neurips.cc/paper_files/paper/2020/file/1457c0d6bfcb4967418bfb8ac142f64a-Paper.pdf} {Language models are few-shot learners}.
\newblock In \emph{Advances in Neural Information Processing Systems}, volume~33, pages 1877--1901. Curran Associates, Inc.

\bibitem[{Cheng et~al.(2021)Cheng, Guo, Shu, and Liu}]{cheng2021causal}
Lu~Cheng, Ruocheng Guo, Kai Shu, and Huan Liu. 2021.
\newblock Causal understanding of fake news dissemination on social media.
\newblock In \emph{KDD}.

\bibitem[{Derczynski et~al.(2017)Derczynski, Bontcheva, Liakata, Procter, Wong Sak~Hoi, and Zubiaga}]{derczynski-etal-2017-semeval}
Leon Derczynski, Kalina Bontcheva, Maria Liakata, Rob Procter, Geraldine Wong Sak~Hoi, and Arkaitz Zubiaga. 2017.
\newblock \href {https://doi.org/10.18653/v1/S17-2006} {{S}em{E}val-2017 task 8: {R}umour{E}val: Determining rumour veracity and support for rumours}.
\newblock In \emph{Proceedings of the 11th International Workshop on Semantic Evaluation ({S}em{E}val-2017)}, pages 69--76, Vancouver, Canada. Association for Computational Linguistics.

\bibitem[{Devlin et~al.(2019)Devlin, Chang, Lee, and Toutanova}]{devlin-etal-2019-bert}
Jacob Devlin, Ming-Wei Chang, Kenton Lee, and Kristina Toutanova. 2019.
\newblock \href {https://doi.org/10.18653/v1/N19-1423} {{BERT}: Pre-training of deep bidirectional transformers for language understanding}.
\newblock In \emph{Proceedings of the 2019 Conference of the North {A}merican Chapter of the Association for Computational Linguistics: Human Language Technologies, Volume 1 (Long and Short Papers)}, pages 4171--4186, Minneapolis, Minnesota. Association for Computational Linguistics.

\bibitem[{Dietterich et~al.(1997)Dietterich, Lathrop, and Lozano-Pérez}]{DIETTERICH199731}
Thomas~G. Dietterich, Richard~H. Lathrop, and Tomás Lozano-Pérez. 1997.
\newblock \href {https://doi.org/https://doi.org/10.1016/S0004-3702(96)00034-3} {Solving the multiple instance problem with axis-parallel rectangles}.
\newblock \emph{Artificial Intelligence}, 89(1):31--71.

\bibitem[{Dungs et~al.(2018)Dungs, Aker, Fuhr, and Bontcheva}]{dungs-etal-2018-rumour}
Sebastian Dungs, Ahmet Aker, Norbert Fuhr, and Kalina Bontcheva. 2018.
\newblock \href {https://aclanthology.org/C18-1284} {Can rumour stance alone predict veracity?}
\newblock In \emph{Proceedings of the 27th International Conference on Computational Linguistics}, pages 3360--3370, Santa Fe, New Mexico, USA. Association for Computational Linguistics.

\bibitem[{Gorrell et~al.(2019)Gorrell, Kochkina, Liakata, Aker, Zubiaga, Bontcheva, and Derczynski}]{gorrell-etal-2019-semeval}
Genevieve Gorrell, Elena Kochkina, Maria Liakata, Ahmet Aker, Arkaitz Zubiaga, Kalina Bontcheva, and Leon Derczynski. 2019.
\newblock \href {https://doi.org/10.18653/v1/S19-2147} {{S}em{E}val-2019 task 7: {R}umour{E}val, determining rumour veracity and support for rumours}.
\newblock In \emph{Proceedings of the 13th International Workshop on Semantic Evaluation}, pages 845--854, Minneapolis, Minnesota, USA. Association for Computational Linguistics.

\bibitem[{Houlsby et~al.(2019)Houlsby, Giurgiu, Jastrzebski, Morrone, De~Laroussilhe, Gesmundo, Attariyan, and Gelly}]{pmlr-v97-houlsby19a}
Neil Houlsby, Andrei Giurgiu, Stanislaw Jastrzebski, Bruna Morrone, Quentin De~Laroussilhe, Andrea Gesmundo, Mona Attariyan, and Sylvain Gelly. 2019.
\newblock \href {https://proceedings.mlr.press/v97/houlsby19a.html} {Parameter-efficient transfer learning for {NLP}}.
\newblock In \emph{Proceedings of the 36th International Conference on Machine Learning}, volume~97 of \emph{Proceedings of Machine Learning Research}, pages 2790--2799. PMLR.

\bibitem[{Hu et~al.(2021)Hu, Wallis, Allen-Zhu, Li, Wang, Wang, Chen et~al.}]{hu2021lora}
Edward~J Hu, Phillip Wallis, Zeyuan Allen-Zhu, Yuanzhi Li, Shean Wang, Lu~Wang, Weizhu Chen, et~al. 2021.
\newblock Lora: Low-rank adaptation of large language models.
\newblock In \emph{International Conference on Learning Representations}.

\bibitem[{James Richard~Foulds(2010)}]{FouldsFrank2010}
Eibe~Frank James Richard~Foulds. 2010.
\newblock \href {https://doi.org/https://doi.org/10.1017/S026988890999035X} {A review of multi-instance learning assumptions}.
\newblock \emph{The Knowledge Engineering Review}, 25(1):1--25.

\bibitem[{Joshi et~al.(2015)Joshi, Kale, Chandel, and Pal}]{joshi2015likert}
Ankur Joshi, Saket Kale, Satish Chandel, and D~Kumar Pal. 2015.
\newblock Likert scale: Explored and explained.
\newblock \emph{British journal of applied science \& technology}, 7(4):396.

\bibitem[{Kanamori(2010)}]{kanamori2010deformation}
Takafumi Kanamori. 2010.
\newblock Deformation of log-likelihood loss function for multiclass boosting.
\newblock \emph{Neural Networks}, 23(7):843--864.

\bibitem[{Khoo et~al.(2020)Khoo, Chieu, Qian, and Jiang}]{khoo2020interpretable}
Ling Min~Serena Khoo, Hai~Leong Chieu, Zhong Qian, and Jing Jiang. 2020.
\newblock Interpretable rumor detection in microblogs by attending to user interactions.
\newblock In \emph{AAAI}.

\bibitem[{Kobbe et~al.(2020)Kobbe, Hulpuș, and Stuckenschmidt}]{kobbe2020unsupervised}
Jonathan Kobbe, Ioana Hulpuș, and Heiner Stuckenschmidt. 2020.
\newblock Unsupervised stance detection for arguments from consequences.
\newblock In \emph{Proceedings of the 2020 Conference on Empirical Methods in Natural Language Processing (EMNLP)}, pages 50--60.

\bibitem[{Kochkina et~al.(2017)Kochkina, Liakata, and Augenstein}]{kochkina2017turing}
Elena Kochkina, Maria Liakata, and Isabelle Augenstein. 2017.
\newblock Turing at semeval-2017 task 8: Sequential approach to rumour stance classification with branch-lstm.
\newblock In \emph{Proceedings of the 11th International Workshop on Semantic Evaluation (SemEval-2017)}, pages 475--480.

\bibitem[{Kochkina et~al.(2018)Kochkina, Liakata, and Zubiaga}]{kochkina2018all}
Elena Kochkina, Maria Liakata, and Arkaitz Zubiaga. 2018.
\newblock All-in-one: Multi-task learning for rumour verification.
\newblock In \emph{Proceedings of the 26th International Conference on Computational Linguistics}, pages 3402--3413.

\bibitem[{K\"{u}\c{c}\"{u}k and Can(2020)}]{10.1145/3369026}
Dilek K\"{u}\c{c}\"{u}k and Fazli Can. 2020.
\newblock \href {https://doi.org/10.1145/3369026} {Stance detection: A survey}.
\newblock \emph{ACM Comput. Surv.}, 53(1).

\bibitem[{Lee et~al.(2021)Lee, Bang, Madotto, and Fung}]{lee-etal-2021-towards}
Nayeon Lee, Yejin Bang, Andrea Madotto, and Pascale Fung. 2021.
\newblock \href {https://doi.org/10.18653/v1/2021.naacl-main.158} {Towards few-shot fact-checking via perplexity}.
\newblock In \emph{Proceedings of the 2021 Conference of the North American Chapter of the Association for Computational Linguistics: Human Language Technologies}, pages 1971--1981, Online. Association for Computational Linguistics.

\bibitem[{Lester et~al.(2021)Lester, Al-Rfou, and Constant}]{lester-etal-2021-power}
Brian Lester, Rami Al-Rfou, and Noah Constant. 2021.
\newblock \href {https://doi.org/10.18653/v1/2021.emnlp-main.243} {The power of scale for parameter-efficient prompt tuning}.
\newblock In \emph{Proceedings of the 2021 Conference on Empirical Methods in Natural Language Processing}, pages 3045--3059, Online and Punta Cana, Dominican Republic. Association for Computational Linguistics.

\bibitem[{Li et~al.(2020)Li, Sujana, and Kao}]{li-etal-2020-exploiting}
Jiawen Li, Yudianto Sujana, and Hung-Yu Kao. 2020.
\newblock \href {https://doi.org/10.18653/v1/2020.coling-main.473} {Exploiting microblog conversation structures to detect rumors}.
\newblock In \emph{Proceedings of the 28th International Conference on Computational Linguistics}, pages 5420--5429, Barcelona, Spain (Online). International Committee on Computational Linguistics.

\bibitem[{Li et~al.(2019)Li, Zhang, and Si}]{li-etal-2019-rumor-detection}
Quanzhi Li, Qiong Zhang, and Luo Si. 2019.
\newblock \href {https://doi.org/10.18653/v1/P19-1113} {Rumor detection by exploiting user credibility information, attention and multi-task learning}.
\newblock In \emph{Proceedings of the 57th Annual Meeting of the Association for Computational Linguistics}, pages 1173--1179, Florence, Italy. Association for Computational Linguistics.

\bibitem[{Li et~al.(2021{\natexlab{a}})Li, Sosea, Sawant, Nair, Inkpen, and Caragea}]{li-etal-2021-p}
Yingjie Li, Tiberiu Sosea, Aditya Sawant, Ajith~Jayaraman Nair, Diana Inkpen, and Cornelia Caragea. 2021{\natexlab{a}}.
\newblock \href {https://doi.org/10.18653/v1/2021.findings-acl.208} {{P}-stance: A large dataset for stance detection in political domain}.
\newblock In \emph{Findings of the Association for Computational Linguistics: ACL-IJCNLP 2021}, pages 2355--2365, Online. Association for Computational Linguistics.

\bibitem[{Li et~al.(2021{\natexlab{b}})Li, Sosea, Sawant, Nair, Inkpen, and Caragea}]{li2021p}
Yingjie Li, Tiberiu Sosea, Aditya Sawant, Ajith~Jayaraman Nair, Diana Inkpen, and Cornelia Caragea. 2021{\natexlab{b}}.
\newblock P-stance: A large dataset for stance detection in political domain.
\newblock In \emph{Findings of the Association for Computational Linguistics: ACL-IJCNLP 2021}, pages 2355--2365.

\bibitem[{Li et~al.(2023)Li, Zhao, and Caragea}]{li2023tts}
Yingjie Li, Chenye Zhao, and Cornelia Caragea. 2023.
\newblock Tts: A target-based teacher-student framework for zero-shot stance detection.
\newblock In \emph{Proceedings of the ACM Web Conference 2023}, pages 1500--1509.

\bibitem[{Liang et~al.(2021)Liang, Fu, Gui, Yang, Du, He, and Xu}]{liang2021target}
Bin Liang, Yonghao Fu, Lin Gui, Min Yang, Jiachen Du, Yulan He, and Ruifeng Xu. 2021.
\newblock Target-adaptive graph for cross-target stance detection.
\newblock In \emph{Proceedings of the Web Conference 2021}, pages 3453--3464.

\bibitem[{Liang et~al.(2022)Liang, Zhu, Li, Yang, Gui, He, and Xu}]{liang2022jointcl}
Bin Liang, Qinglin Zhu, Xiang Li, Min Yang, Lin Gui, Yulan He, and Ruifeng Xu. 2022.
\newblock Jointcl: A joint contrastive learning framework for zero-shot stance detection.
\newblock In \emph{Proceedings of the 60th Annual Meeting of the Association for Computational Linguistics (Volume 1: Long Papers)}, pages 81--91.

\bibitem[{Lin et~al.(2022)Lin, Ma, Chen, Yang, Cheng, and Guang}]{lin2022detect}
Hongzhan Lin, Jing Ma, Liangliang Chen, Zhiwei Yang, Mingfei Cheng, and Chen Guang. 2022.
\newblock Detect rumors in microblog posts for low-resource domains via adversarial contrastive learning.
\newblock In \emph{Findings of the Association for Computational Linguistics: NAACL 2022}, pages 2543--2556.

\bibitem[{Lin et~al.(2021)Lin, Ma, Cheng, Yang, Chen, and Chen}]{lin2021rumor}
Hongzhan Lin, Jing Ma, Mingfei Cheng, Zhiwei Yang, Liangliang Chen, and Guang Chen. 2021.
\newblock Rumor detection on twitter with claim-guided hierarchical graph attention networks.
\newblock In \emph{Proceedings of the 2021 Conference on Empirical Methods in Natural Language Processing}, pages 10035--10047.

\bibitem[{Lin et~al.(2023)Lin, Yi, Ma, Jiang, Luo, Shi, and Liu}]{lin2023zero}
Hongzhan Lin, Pengyao Yi, Jing Ma, Haiyun Jiang, Ziyang Luo, Shuming Shi, and Ruifang Liu. 2023.
\newblock Zero-shot rumor detection with propagation structure via prompt learning.
\newblock In \emph{Proceedings of the AAAI Conference on Artificial Intelligence}, volume~37, pages 5213--5221.

\bibitem[{Liu et~al.(2015)Liu, Nourbakhsh, Li, Fang, and Shah}]{liu2015real}
Xiaomo Liu, Armineh Nourbakhsh, Quanzhi Li, Rui Fang, and Sameena Shah. 2015.
\newblock Real-time rumor debunking on twitter.
\newblock In \emph{Proceedings of the 24th ACM international on conference on information and knowledge management}, pages 1867--1870.

\bibitem[{Liu et~al.(2019)Liu, Ott, Goyal, Du, Joshi, Chen, Levy, Lewis, Zettlemoyer, and Stoyanov}]{liu2019roberta}
Yinhan Liu, Myle Ott, Naman Goyal, Jingfei Du, Mandar Joshi, Danqi Chen, Omer Levy, Mike Lewis, Luke Zettlemoyer, and Veselin Stoyanov. 2019.
\newblock Roberta: A robustly optimized bert pretraining approach.
\newblock \emph{arXiv preprint arXiv:1907.11692}.

\bibitem[{Lu and Li(2020)}]{lu2020gcan}
Yi-Ju Lu and Cheng-Te Li. 2020.
\newblock Gcan: Graph-aware co-attention networks for explainable fake news detection on social media.
\newblock In \emph{ACL}.

\bibitem[{Lukasik et~al.(2016)Lukasik, Srijith, Vu, Bontcheva, Zubiaga, and Cohn}]{lukasik2016hawkes}
Michal Lukasik, PK~Srijith, Duy Vu, Kalina Bontcheva, Arkaitz Zubiaga, and Trevor Cohn. 2016.
\newblock Hawkes processes for continuous time sequence classification: an application to rumour stance classification in twitter.
\newblock In \emph{Proceedings of the 54th Annual Meeting of the Association for Computational Linguistics (Volume 2: Short Papers)}, pages 393--398.

\bibitem[{Ma and Gao(2020)}]{ma2020debunking}
Jing Ma and Wei Gao. 2020.
\newblock Debunking rumors on twitter with tree transformer.
\newblock In \emph{Proceedings of the 28th International Conference on Computational Linguistics}, pages 5455--5466.

\bibitem[{Ma et~al.(2020)Ma, Gao, Joty, and Wong}]{ma2020attention}
Jing Ma, Wei Gao, Shafiq Joty, and Kam-Fai Wong. 2020.
\newblock An attention-based rumor detection model with tree-structured recursive neural networks.
\newblock \emph{ACM Transactions on Intelligent Systems and Technology (TIST)}, 11(4):1--28.

\bibitem[{Ma et~al.(2016)Ma, Gao, Mitra, Kwon, Jansen, Wong, and Cha}]{ma2016detecting}
Jing Ma, Wei Gao, Prasenjit Mitra, Sejeong Kwon, Bernard~J Jansen, Kam-Fai Wong, and Meeyoung Cha. 2016.
\newblock Detecting rumors from microblogs with recurrent neural networks.
\newblock In \emph{IJCAI}.

\bibitem[{Ma et~al.(2017)Ma, Gao, and Wong}]{ma2017detect}
Jing Ma, Wei Gao, and Kam-Fai Wong. 2017.
\newblock Detect rumors in microblog posts using propagation structure via kernel learning.
\newblock In \emph{Proceedings of the 55th Annual Meeting of the Association for Computational Linguistics (Volume 1: Long Papers)}, pages 708--717.

\bibitem[{Ma et~al.(2018)Ma, Gao, and Wong}]{ma2018detect}
Jing Ma, Wei Gao, and Kam-Fai Wong. 2018.
\newblock Detect rumor and stance jointly by neural multi-task learning.
\newblock In \emph{Companion proceedings of the the web conference 2018}, pages 585--593.

\bibitem[{Ma et~al.(2021)Ma, Li, Gao, Yang, and Wong}]{ma2021improving}
Jing Ma, Jun Li, Wei Gao, Yang Yang, and Kam-Fai Wong. 2021.
\newblock Improving rumor detection by promoting information campaigns with transformer-based generative adversarial learning.
\newblock \emph{IEEE Transactions on Knowledge and Data Engineering}.

\bibitem[{Nguyen et~al.(2020)Nguyen, Vu, and Nguyen}]{nguyen2020bertweet}
Dat~Quoc Nguyen, Thanh Vu, and Anh~Tuan Nguyen. 2020.
\newblock Bertweet: A pre-trained language model for english tweets.
\newblock In \emph{EMNLP}.

\bibitem[{Pan et~al.(2023)Pan, Wu, Lu, Luu, Wang, Kan, and Nakov}]{pan-etal-2023-fact}
Liangming Pan, Xiaobao Wu, Xinyuan Lu, Anh~Tuan Luu, William~Yang Wang, Min-Yen Kan, and Preslav Nakov. 2023.
\newblock \href {https://doi.org/10.18653/v1/2023.acl-long.386} {Fact-checking complex claims with program-guided reasoning}.
\newblock In \emph{Proceedings of the 61st Annual Meeting of the Association for Computational Linguistics (Volume 1: Long Papers)}, pages 6981--7004, Toronto, Canada. Association for Computational Linguistics.

\bibitem[{Pick et~al.(2022)Pick, Kozhukhov, Vilenchik, and Tsur}]{pick2022stem}
Ron~Korenblum Pick, Vladyslav Kozhukhov, Dan Vilenchik, and Oren Tsur. 2022.
\newblock Stem: unsupervised structural embedding for stance detection.
\newblock In \emph{Proceedings of the AAAI Conference on Artificial Intelligence}, volume~36, pages 11174--11182.

\bibitem[{Qazvinian et~al.(2011)Qazvinian, Rosengren, Radev, and Mei}]{qazvinian-etal-2011-rumor}
Vahed Qazvinian, Emily Rosengren, Dragomir~R. Radev, and Qiaozhu Mei. 2011.
\newblock \href {https://aclanthology.org/D11-1147} {Rumor has it: Identifying misinformation in microblogs}.
\newblock In \emph{Proceedings of the 2011 Conference on Empirical Methods in Natural Language Processing}, pages 1589--1599, Edinburgh, Scotland, UK. Association for Computational Linguistics.

\bibitem[{Radford et~al.(2019)Radford, Wu, Child, Luan, Amodei, Sutskever et~al.}]{radford2019language}
Alec Radford, Jeffrey Wu, Rewon Child, David Luan, Dario Amodei, Ilya Sutskever, et~al. 2019.
\newblock Language models are unsupervised multitask learners.
\newblock \emph{OpenAI blog}, 1(8):9.

\bibitem[{Ran and Jia(2023)}]{DBLP:conf/aaai/RanJ23}
Hongyan Ran and Caiyan Jia. 2023.
\newblock \href {https://doi.org/10.1609/aaai.v37i11.26584} {Unsupervised cross-domain rumor detection with contrastive learning and cross-attention}.
\newblock In \emph{Thirty-Seventh {AAAI} Conference on Artificial Intelligence, {AAAI} 2023}, pages 13510--13518. {AAAI} Press.

\bibitem[{Roozenbeek and van~der Linden(2019)}]{roozenbeek2019fake}
Jon Roozenbeek and Sander van~der Linden. 2019.
\newblock Fake news game confers psychological resistance against online misinformation.
\newblock \emph{Palgrave Communications}, 5(1):1--10.

\bibitem[{Rosenfeld et~al.(2020)Rosenfeld, Szanto, and Parkes}]{rosenfeld2020kernel}
Nir Rosenfeld, Aron Szanto, and David~C Parkes. 2020.
\newblock A kernel of truth: Determining rumor veracity on twitter by diffusion pattern alone.
\newblock In \emph{Proceedings of The Web Conference 2020}, pages 1018--1028.

\bibitem[{Sun et~al.(2022)Sun, Zhang, Zheng, and Ma}]{sun2022ddgcn}
Mengzhu Sun, Xi~Zhang, Jiaqi Zheng, and Guixiang Ma. 2022.
\newblock Ddgcn: Dual dynamic graph convolutional networks for rumor detection on social media.
\newblock In \emph{Proceedings of the AAAI conference on artificial intelligence}, volume~36, pages 4611--4619.

\bibitem[{Sutton and Barto(2018)}]{sutton2018reinforcement}
Richard~S Sutton and Andrew~G Barto. 2018.
\newblock \emph{Reinforcement learning: An introduction}.
\newblock MIT press.

\bibitem[{Taori et~al.(2023)Taori, Gulrajani, Zhang, Dubois, Li, Guestrin, Liang, and Hashimoto}]{taori2023alpaca}
Rohan Taori, Ishaan Gulrajani, Tianyi Zhang, Yann Dubois, Xuechen Li, Carlos Guestrin, Percy Liang, and Tatsunori~B Hashimoto. 2023.
\newblock Alpaca: A strong, replicable instruction-following model.
\newblock \emph{Stanford Center for Research on Foundation Models. https://crfm. stanford. edu/2023/03/13/alpaca. html}, 3(6):7.

\bibitem[{Tian et~al.(2020)Tian, Zhang, Wang, and Liu}]{10.1007/978-3-030-45439-5_38}
Lin Tian, Xiuzhen Zhang, Yan Wang, and Huan Liu. 2020.
\newblock Early detection of rumours on twitter via stance transfer learning.
\newblock In \emph{Advances in Information Retrieval}, pages 575--588, Cham. Springer International Publishing.

\bibitem[{Touvron et~al.(2023)Touvron, Martin, Stone, Albert, Almahairi, Babaei, Bashlykov, Batra, Bhargava, Bhosale et~al.}]{touvron2023Llama}
Hugo Touvron, Louis Martin, Kevin Stone, Peter Albert, Amjad Almahairi, Yasmine Babaei, Nikolay Bashlykov, Soumya Batra, Prajjwal Bhargava, Shruti Bhosale, et~al. 2023.
\newblock Llama 2: Open foundation and fine-tuned chat models.
\newblock \emph{arXiv preprint arXiv:2307.09288}.

\bibitem[{Van~der Maaten and Hinton(2008)}]{van2008visualizing}
Laurens Van~der Maaten and Geoffrey Hinton. 2008.
\newblock Visualizing data using t-sne.
\newblock \emph{Journal of machine learning research}, 9(11).

\bibitem[{Vosoughi et~al.(2018)Vosoughi, Roy, and Aral}]{vosoughi2018spread}
Soroush Vosoughi, Deb Roy, and Sinan Aral. 2018.
\newblock The spread of true and false news online.
\newblock \emph{Science}, 359(6380):1146--1151.

\bibitem[{Wei et~al.(2019{\natexlab{a}})Wei, Mao, and Chen}]{wei2019topic}
Penghui Wei, Wenji Mao, and Guandan Chen. 2019{\natexlab{a}}.
\newblock A topic-aware reinforced model for weakly supervised stance detection.
\newblock In \emph{Proceedings of the aaai conference on artificial intelligence}, volume~33, pages 7249--7256.

\bibitem[{Wei et~al.(2019{\natexlab{b}})Wei, Xu, and Mao}]{wei2019modeling}
Penghui Wei, Nan Xu, and Wenji Mao. 2019{\natexlab{b}}.
\newblock Modeling conversation structure and temporal dynamics for jointly predicting rumor stance and veracity.
\newblock In \emph{Proceedings of the 2019 Conference on Empirical Methods in Natural Language Processing and the 9th International Joint Conference on Natural Language Processing (EMNLP-IJCNLP)}, pages 4787--4798.

\bibitem[{Yang et~al.(2012)Yang, Liu, Yu, and Yang}]{yang2012automatic}
Fan Yang, Yang Liu, Xiaohui Yu, and Min Yang. 2012.
\newblock Automatic detection of rumor on sina weibo.
\newblock In \emph{KDD}.

\bibitem[{Yang et~al.(2022)Yang, Ma, Lin, and Gao}]{yang2022weakly}
Ruichao Yang, Jing Ma, Hongzhan Lin, and Wei Gao. 2022.
\newblock A weakly supervised propagation model for rumor verification and stance detection with multiple instance learning.
\newblock In \emph{Proceedings of the 45th International ACM SIGIR Conference on Research and Development in Information Retrieval}, pages 1761--1772.

\bibitem[{Yu et~al.(2017)Yu, Liu, Wu, Wang, and Tan}]{yu2017convolutional}
Feng Yu, Qiang Liu, Shu Wu, Liang Wang, and Tieniu Tan. 2017.
\newblock A convolutional approach for misinformation identification.
\newblock In \emph{Proceedings of IJCAI}, pages 3901--3907.

\bibitem[{Yu et~al.(2020)Yu, Jiang, Khoo, Chieu, and Xia}]{yu-etal-2020-coupled}
Jianfei Yu, Jing Jiang, Ling Min~Serena Khoo, Hai~Leong Chieu, and Rui Xia. 2020.
\newblock \href {https://doi.org/10.18653/v1/2020.emnlp-main.108} {Coupled hierarchical transformer for stance-aware rumor verification in social media conversations}.
\newblock In \emph{Proceedings of the 2020 Conference on Empirical Methods in Natural Language Processing (EMNLP)}, pages 1392--1401, Online. Association for Computational Linguistics.

\bibitem[{Yuan et~al.(2021)Yuan, Qian, Ma, Zhou, and Hu}]{yuan2021srlf}
Chunyuan Yuan, Wanhui Qian, Qianwen Ma, Wei Zhou, and Songlin Hu. 2021.
\newblock Srlf: a stance-aware reinforcement learning framework for content-based rumor detection on social media.
\newblock In \emph{2021 International Joint Conference on Neural Networks (IJCNN)}, pages 1--8. IEEE.

\bibitem[{Zeng and Gao(2023)}]{zeng-gao-2023-prompt}
Fengzhu Zeng and Wei Gao. 2023.
\newblock \href {https://doi.org/10.18653/v1/2023.findings-acl.278} {Prompt to be consistent is better than self-consistent? few-shot and zero-shot fact verification with pre-trained language models}.
\newblock In \emph{Findings of the Association for Computational Linguistics: ACL 2023}, pages 4555--4569, Toronto, Canada. Association for Computational Linguistics.

\bibitem[{Zhang et~al.(2019)Zhang, Liang, Lipani, Ren, and Yilmaz}]{zhang2019stances}
Qiang Zhang, Shangsong Liang, Aldo Lipani, Zhaochun Ren, and Emine Yilmaz. 2019.
\newblock From stances' imbalance to their hierarchical representation and detection.
\newblock In \emph{The World Wide Web Conference}, pages 2323--2332.

\bibitem[{Zhang and Gao(2023)}]{zhang-gao:2023:ijcnlp}
Xuan Zhang and Wei Gao. 2023.
\newblock \href {https://aclanthology.org/2023.ijcnlp-long.64} {Towards llm-based fact verification on news claims with a hierarchical step-by-step prompting method}.
\newblock In \emph{Proceedings of the 13th International Joint Conference on Natural Language Processing and the 3rd Conference of the Asia-Pacific Chapter of the Association for Computational Linguistics}, pages 996--1011, Nusa Dua, Bali. Association for Computational Linguistics.

\bibitem[{Zhang and Gao(2024)}]{zhang-gao-2024-reinforcement-retrieval}
Xuan Zhang and Wei Gao. 2024.
\newblock \href {https://aclanthology.org/2024.lrec-main.1209} {Reinforcement retrieval leveraging fine-grained feedback for fact checking news claims with black-box {LLM}}.
\newblock In \emph{Proceedings of the 2024 Joint International Conference on Computational Linguistics, Language Resources and Evaluation (LREC-COLING 2024)}, pages 13861--13873, Torino, Italia. ELRA and ICCL.

\bibitem[{Zhao et~al.(2015)Zhao, Resnick, and Mei}]{zhao2015enquiring}
Zhe Zhao, Paul Resnick, and Qiaozhu Mei. 2015.
\newblock Enquiring minds: Early detection of rumors in social media from enquiry posts.
\newblock In \emph{Proceedings of the 24th international conference on world wide web}, pages 1395--1405.

\bibitem[{Zubiaga et~al.(2016{\natexlab{a}})Zubiaga, Kochkina, Liakata, Procter, and Lukasik}]{zubiaga2016stance}
Arkaitz Zubiaga, Elena Kochkina, Maria Liakata, Rob Procter, and Michal Lukasik. 2016{\natexlab{a}}.
\newblock Stance classification in rumours as a sequential task exploiting the tree structure of social media conversations.
\newblock In \emph{Proceedings of COLING 2016, the 26th International Conference on Computational Linguistics: Technical Papers}, pages 2438--2448.

\bibitem[{Zubiaga et~al.(2018)Zubiaga, Kochkina, Liakata, Procter, Lukasik, Bontcheva, Cohn, and Augenstein}]{zubiaga2018discourse}
Arkaitz Zubiaga, Elena Kochkina, Maria Liakata, Rob Procter, Michal Lukasik, Kalina Bontcheva, Trevor Cohn, and Isabelle Augenstein. 2018.
\newblock Discourse-aware rumour stance classification in social media using sequential classifiers.
\newblock \emph{Information Processing \& Management}, 54(2):273--290.

\bibitem[{Zubiaga et~al.(2016{\natexlab{b}})Zubiaga, Liakata, Procter, Wong Sak~Hoi, and Tolmie}]{10.1371/journal.pone.0150989}
Arkaitz Zubiaga, Maria Liakata, Rob Procter, Geraldine Wong Sak~Hoi, and Peter Tolmie. 2016{\natexlab{b}}.
\newblock \href {https://doi.org/10.1371/journal.pone.0150989} {Analysing how people orient to and spread rumours in social media by looking at conversational threads}.
\newblock \emph{PLOS ONE}, 11(3):1--29.

\bibitem[{Zubiaga et~al.(2017)Zubiaga, Voss, Procter, Liakata, Wang, and Tsakalidis}]{zubiaga2017towards}
Arkaitz Zubiaga, Alex Voss, Rob Procter, Maria Liakata, Bo~Wang, and Adam Tsakalidis. 2017.
\newblock Towards real-time, country-level location classification of worldwide tweets.
\newblock \emph{IEEE Transactions on Knowledge and Data Engineering}, 29(9):2053--2066.

\end{thebibliography}

\appendix

\section{Prompt Design} \label{app:prompt}

\subsection{Stance Detection Prompt}\label{sect:SDprompt}

For stance labeling, we use the following prompt: 

{\textit{\small{There is a claim [CLAIM] and I will give you its corresponding conversation thread on Twitter. The conversation thread consists of a sequence of posts and each post [POST] is written by a user [USER]. Please decide the stance expressed by each post towards the claim and explain why, and the stance can be one of the following labels: Support, Deny, Question, and Comment. Please follow the format: Stance: [stance], Reason:[reason].}}}

\begin{figure*}[hbtb]
  \centering
  \includegraphics[width=5.6in]{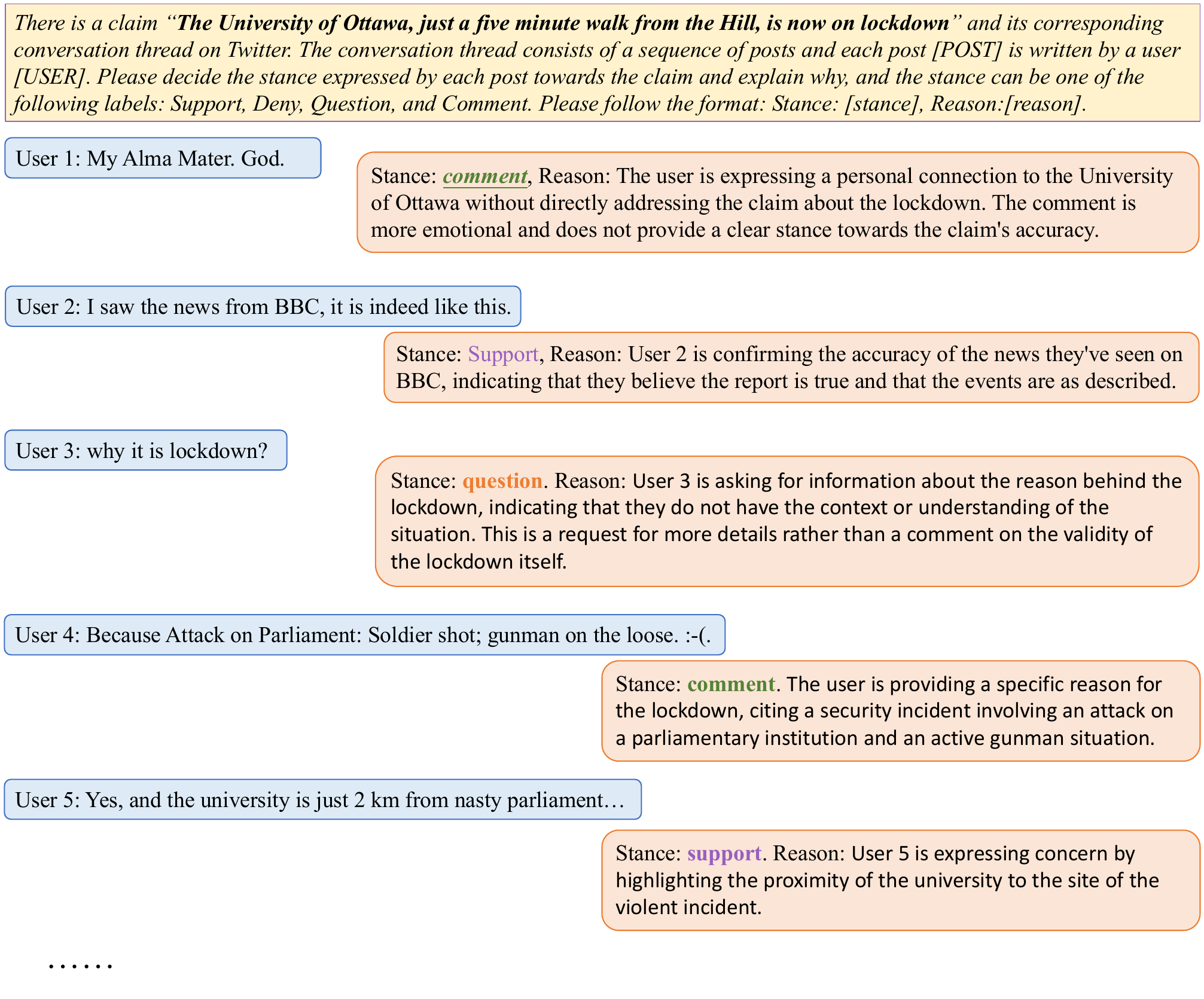}
  \caption{An example of stance prompt.} 
  \label{fig:StancePrompt}
\end{figure*}

The SD LLM will output a stance label and an explanation of labeling rationale for each input post.
We show the details of stance detection prompt in Figure~\ref{fig:StancePrompt}. The blue box represents all the context, the text in the orange box represents the stance and reason output from the stance LLM.

\subsection{Rumor Verification Prompt}\label{sect:RVprompt}

Given a claim and its related posts that are retained by the selection policy model, RV LLM network is prompted to generate veracity label for the claim and a brief explanation of the decision, considering the claim and posts content and stance. The prompt is constructed as follows:

\textit{\small{There is a claim [CLAIM], I will give you its related posts, each expressing a stance toward this claim. Please determine the veracity of the claim, categorizing it as 'True Rumor,' 'False Rumor,' 'Unverified Rumor,' or 'Non-Rumor,' and explain your reasoning. Please follows the format: Veracity: [veracity], Reason: [reason].}}

\begin{figure*}[hbtb]
  \centering
  \includegraphics[width=5.6in]{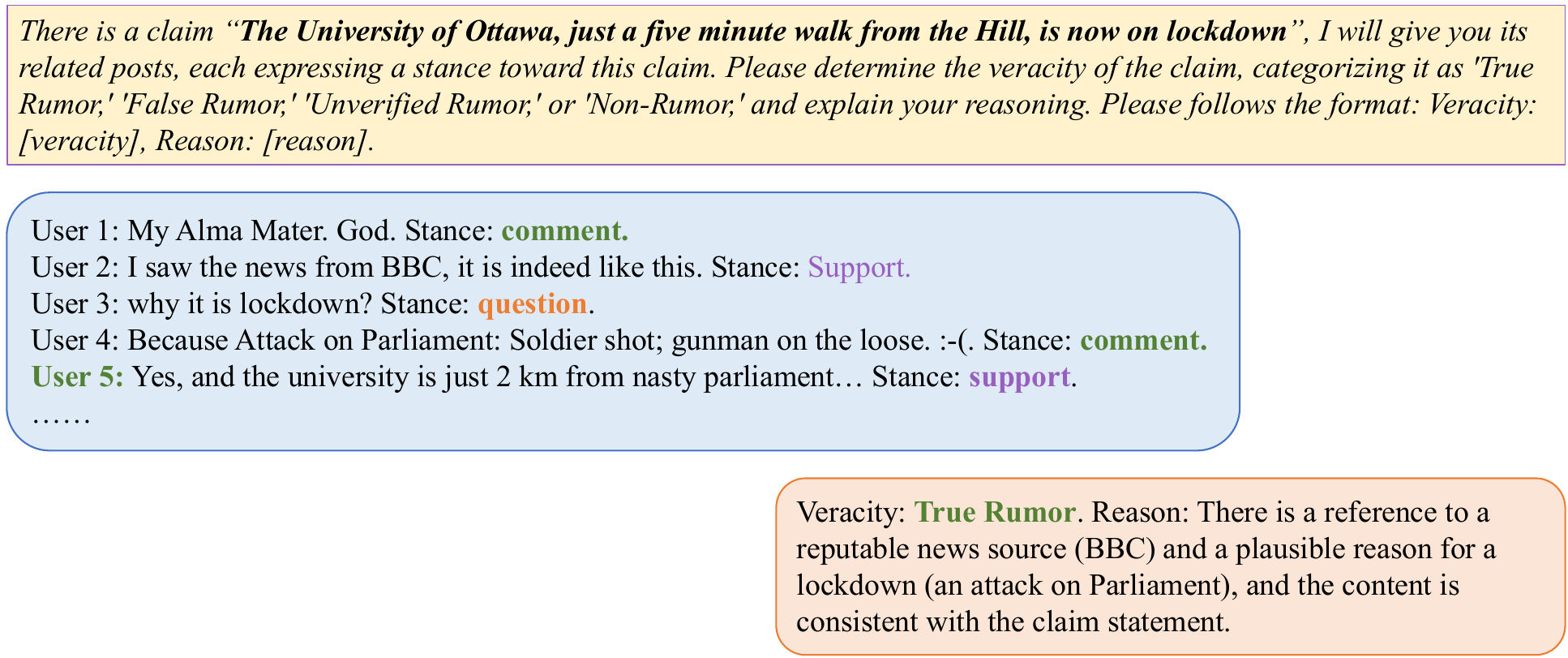}
  \caption{An example of rumor prompt.} 
  \label{fig:RumorPrompt}
\end{figure*}

The RV LLM will output a veracity label and an explanation of labeling rationale for any input claim. We show the details of claim veracity classification prompt in Figure~\ref{fig:RumorPrompt}. The blue box represents all the context, the text in the orange box represents the outputs of claim veracity and the reason.

\section{Dataset Statistics}\label{sect:datastat}
We provide the statistics of training and testing dataset in Table~\ref{tab:TrainData} and Table~\ref{tab:TestData}, respectively.
\begin{table}[h!]
  \centering
  \setlength{\abovecaptionskip}{0.08cm} 
  \resizebox{0.48\textwidth}{!}{
    \begin{tabular}{l|ccc}
    \toprule
    Statistics & Twitter15 & Twitter16  & PHEME \\
    \midrule
    Total claims & 1,308  & 818   & 6,425  \\
    Non-rumor & 374 (28.6\%) & 205 (25.1\%) & 4,023 (62.6\%)  \\
    False-rumor & 370 (28.3\%) & 207 (25.3\%) & 638 (9.9\%)  \\
    True-rumor & 190 (14.5\%) & 205 (25.1\%) & 1,067 (16.6\%) \\
    Unverified-rumor & 374 (28.6\%) & 201 (24.5\%) & 697 (10.8\%) \\
    \midrule
    Total posts & 68,026 & 40,867 & 383,569 \\
    Avg. posts/claim & 52 & 50  & 6 \\
    Max. posts/claim & 814 & 757 & 228 \\
    Min. posts/claim & 1 & 1 & 3 \\
    \bottomrule
    \end{tabular}%
    }
\caption{Statistics of datasets used for training.}
\label{tab:TrainData}
\end{table}%

\begin{table}[h!]
  \centering
  \small
    \begin{tabular}{l|cc}
    \toprule
    Statistics & RumorEval-S & SemEval-8 \\
    \midrule
    Total claims & 425 & 297 \\
    Non-rumor & 100 (23.53\%) & —— \\
    False-rumor & 74 (17.41\%) & 62 (20.8\%) \\
    True-rumor & 145 (34.12\%) & 137 (46.1\%) \\
    Unverified-rumor & 106 (24.94\%) & 98 (33.0\%)\\
    \midrule 			
    Posts of Support & 1,320 (19.65\%) & 910 (20.1\%)\\
    Posts of Deny & 522 (7.77\%) & 334 (7.6\%)\\
    Posts of Question & 531 (7.90\%) & 358 (7.9\%) \\
    Posts of Comment & 4,345 (64.68\%) & 2,907 (64.3\%) \\
    \midrule
    Total posts & 6,718 & 4,519 \\
    Avg. posts/claim & 16 & 15\\
    Max. posts/claim & 249 & 228\\
    Min. posts/claim & 2 & 3\\ 
    \bottomrule
    \end{tabular}%
  \caption{Statistics of the datasets used for testing.}    
  \label{tab:TestData}%
\end{table}%

\section{Hyper-parameter Setting} \label{app:LlamaArgs}
As for Llama 2 model, We download the version with 7 billion parameters(Llama-2-7b) on August, 20, 2023 to obtain its responses. we then employ HuggingFace AutoTokenizer\footnote{\url{https://huggingface.co/docs/transformers/v4.33.0/en/model\_doc/auto\#transformers.AutoTokenizer}} and classes to run the model. The following arguments are setup during fine-tuning stage:

\indent ``model\_name": ``Llama-2-7b"; \\
\indent ``learning\_rate": 1e-4;\\
\indent``num\_train\_epochs": 6; \\
\indent``max\_seq\_length": 4096; \\
\indent``load\_in\_4bit": True; \\
\indent``lr\_scheduler\_type": ``linear"; \\
\indent``temperature": 0

As for selector policy model, we set the hyper-parameters as the following:

\indent ``learning rate": 5e-5 \\
\indent ``optimizer": Adam \\
\indent ``warm-up rate": 0.1 \\
\indent ``batch size": 4 \\
\indent ``maximum epoch": 50


\section{Details of Baseline Models}

\subsection{Stance Detection Baselines} \label{app:stanceBase}
Since our SD network only requires claim veracity label but not post stance label, we choose unsupervised, supervised, and weakly supervised baselines:
(1) \textbf{TGA}~\cite{allaway2020zero}: A topic-grouped attention network 
for zero-shot stance detection.  
(2) \textbf{BerTweet}~\cite{nguyen2020bertweet}: A language model pre-trained on 850M tweets, which is fine-tuned on validataion dataset to adapt to stance detection task.
(3) \textbf{Llama 2-ST}~\cite{touvron2023Llama}: A pre-trained large language model developed by Meta for only stance detection task.
(4) \textbf{Llama 2-MT}~\cite{touvron2023Llama}: A pre-trained large language model prompted to perform multi-task of stance detection and rumor verification together.
(5) \textbf{BiGRU}~\cite{augenstein2016stance}: A bidirectional GRU-based stance detection model. 
(6) \textbf{BrLSTM}~\cite{kochkina2017turing}: An LSTM-based model that models the structured conversational thread to detect stance. 
(7) \textbf{MT-GRU}~\cite{ma2018detect}: A RNN-based multi-task learning model to jointly detect rumors and stances. 
(8) \textbf{JointCL}~\cite{liang2022jointcl}: A zero-shot stance detection model based on contrastive learning.
(9) \textbf{SRLF}~\cite{yuan2021srlf}: A stance-aware reinforced framework for stance detection.\footnote{The original method just outputs rumor veracity result, we put a Softmax layer behind stance embeddings to get stance output here.}
(10) \textbf{TD-MIL}~\cite{yang2022weakly}: A weakly supervised model for stance detection and rumor verification based on top-down tree structure. Additionally,
\textbf{JSDRV (\textsc{DataSet})} is our LLM-based reinforcement tuning method applied for stance detection. 

\subsection{Rumor Verification Baselines} \label{app:rumorBase}

We collect unsupervised, supervised, weakly supervised baselines for rumor verification. Since multi-task learning also can enhance rumor detection, we also introduce two multi-task baselines:
(1) \textbf{BerTweet}~\cite{nguyen2020bertweet}: A pre-trained language model with 850M tweets, and we fine-tune it for rumor verification here. 
(2) \textbf{Llama 2-ST}~\cite{touvron2023Llama}: A large language model pre-trained on various domains myriad corpus, we use it for single rumor verification task.
(3) \textbf{Llama 2-MT}~\cite{touvron2023Llama}: A pre-trained large language model prompted to perform multi-task of stance detection and rumor verification together.
(4) \textbf{GCAN}~\cite{lu2020gcan}: A graph-aware co-attention model utilizing retweet to detect rumor veracity. 
(5) \textbf{TD-RvNN}~\cite{ma2020attention}: A top-down tree-structured attention networks for rumor verification.
(6) \textbf{PLAN}~\cite{khoo2020interpretable}: A transformer based rumor verification model utilizing interactions between users.
(7) \textbf{DDGCN}~\cite{sun2022ddgcn}: A rumor verification model utilizing the dynamic  propagation texts and external knowledge.
(8) \textbf{SRLF}~\cite{yuan2021srlf}: A stance-aware refinforced framework for rumor verification.
(9) \textbf{TD-MIL}~\cite{yang2022weakly}: A top-down tree propagation model for joint detection tasks on stance type and rumor veracity.
(10) \textbf{MTL2}~\cite{kochkina2018all}: A sequential model utilizing a number of task-specific layers for stance detection and rumor verification.
(11) \textbf{MT-GRU}~\cite{ma2018detect}: A multi-task learning approach to jointly detect rumors and stances by capturing both shared and task-specific features. Additionally, 
\textbf{JSDRV(\textsc{DataSet})} is our reinforcement tuning method applied for rumor verification.


\section{Analysis}\label{app:analysis}

\subsection{Case Study}\label{sect:case}
\begin{figure}[t!]
\centering
\includegraphics[scale=0.38]{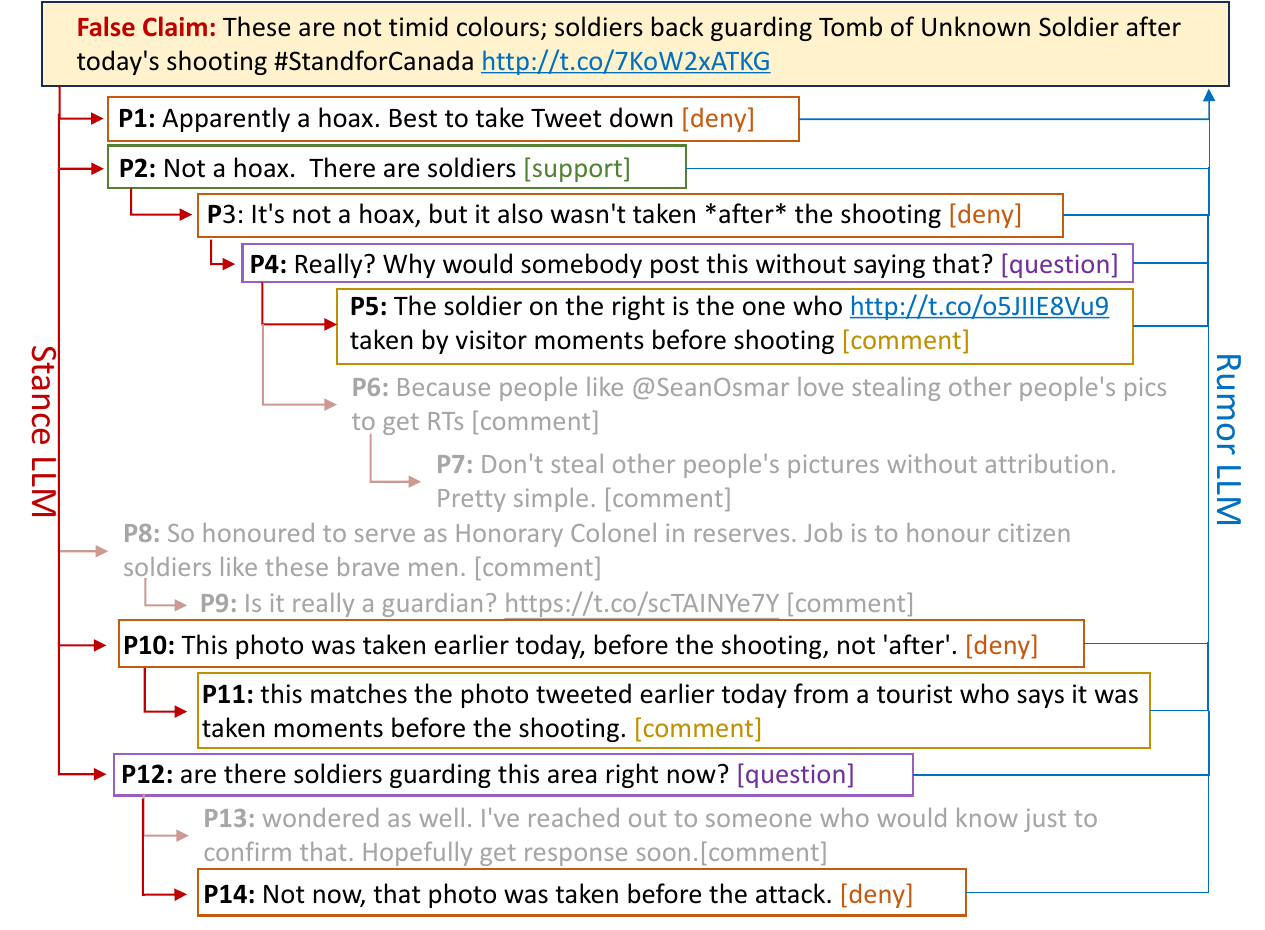}
\label{fig:GlobalAtt}
\caption{Case study. SD LLM predicts posts stance (left red line). The selector policy chooses posts stances (color boxes) for RV LLM to predict rumor veracity (right blue line).}\label{fig:CaseStudy}
\end{figure}

We also plot example outputs of JSDRV in Figure~\ref{fig:CaseStudy}, where the red line represents the stance detection process, the gray texts denote discarded posts, and the blue line indicates the rumor verification process during inference. We observe that the selector policy can choose posts with high-quality stances annotated by stance LLM, which provides useful 
cues to rumor verification LLM.

\subsection{Sensitivity Study}\label{sect:seed}

\begin{figure*}[hbtb]
  \centering
  \includegraphics[width=5.6in]{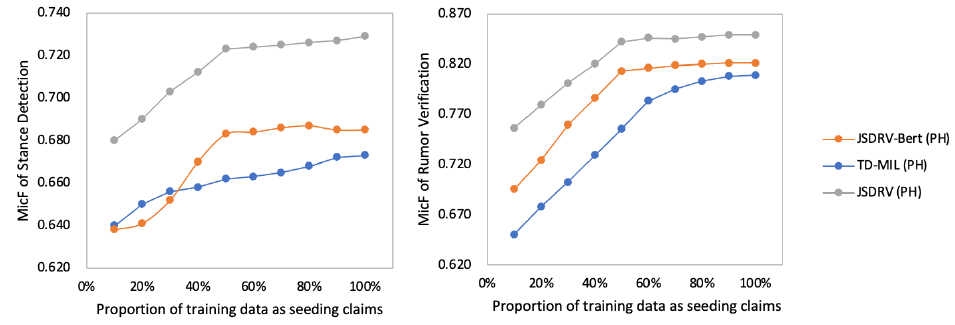}
  \caption{Impact of the size of seeding claims.} 
  \label{fig:SeedClaimsAna}
\end{figure*}

We conduct a sensitivity study to see the impact of the size of seeding claims. Using JSDRV (PH) model, we show the variation of micF score on stance detection and rumor verification tasks with the increase of the proportion of seeding claims. We also show the performance of the strongest baseline TD-MIL trained on the same sets of seeding claims. Figure~\ref{fig:SeedClaimsAna} indicates that (1) With more seeding claims, the performance of TD-MIL (PH), JSDRV (PH), and RSDRV-Bert (PH) all gets improved; (2) The performance of JSDRV tends to stabilize more quickly than TD-MIL. When the ratio reaches 50\%, the performance of JSDRV becomes saturated at a much higher level, while TD-MIL cannot catch up even using up all the training data. (3) JSDRV performs much better than TD-MIL using a same proportion of training data.

\begin{figure*}[hbtb]
  \centering
  \includegraphics[width=5.6in]{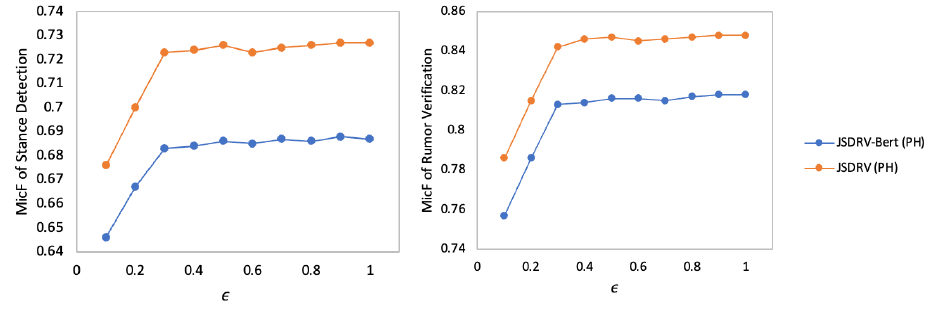}
  \caption{Impact of $\epsilon$.} 
  \label{fig:GreedyControllerAna}
\end{figure*}

To get an intuitive understanding of the greedy controller, we conduct a study to assess the performance of JSDRV-Bert (PH) and JSDRV (PH) across varying $\epsilon$ rates. We use the same $\epsilon$ to control both post and claim  sampling. The MicF scores for stance detection and rumor verification on the RumorEval-S dataset are shown in Figure~\ref{fig:GreedyControllerAna}. Our observations include: (1) The performance of both models improves as $\epsilon$ increases. (2) Model performance stabilizes as $\epsilon$ reaches 0.3, indicating that our greedy controller can achieve comparable results with limited instances labels compared to abundant  instances labels.

\subsection{User Study} \label{sect:UserStudy}
We conduct a user study to evaluate the quality of the model output. We sample 120 samples from RumorEval-S and present them in two forms: Baseline (claim, posts) and JSDRV (claim, selected post-stance pairs, reasons). We then ask 6 users to label the articles and give their confidence in a 5-point Likert Scale~\cite{joshi2015likert}, and each person is given only one form to avoid cross influence. 

 \begin{table}[htbp]
   \centering
   \small
  \resizebox{0.48\textwidth}{!}{
     \begin{tabular}{lcccc}
     \toprule
     & \textbf{F1}    & \textbf{Acc}   & \textbf{Confidence} & \textbf{Avg. Time/news} \\
     \midrule
     \textbf{Baseline} & 0.693 & 0.713 & 1.017 & 25 sec \\
     \textbf{JSDRV} & 0.961 & 0.990 & 4.165 & 5 sec \\
     \bottomrule
     \end{tabular}%
     }
   \caption{User study results.}
   \label{tab:UserStudy}
 \end{table}

Table~\ref{tab:UserStudy} shows that 1) users determine the rumors more accurately with JSDRV, 2) users spent 75\% less time identifying rumors, and 3) users show higher confidence with the results of JSDRV, suggesting that users tend to be more sure about their decision when stance and related reasons have been provided.


\end{document}